%% file: cas-dc-sample.tex
\documentclass[a4paper,fleqn]{cas-dc}
\usepackage[numbers]{natbib}
\usepackage{amsmath,amssymb,amsfonts}
\usepackage{algorithmic}
\usepackage{graphicx}
\usepackage{tabularray}
\usepackage{textcomp}
\usepackage{makecell}
\UseTblrLibrary{booktabs}
\usepackage{subcaption}
\usepackage{multirow}
\usepackage[ruled,vlined]{algorithm2e}
\usepackage{amsmath}
\usepackage{mathtools}
\usepackage{ulem}
\usepackage[table]{xcolor}

\usepackage[most]{tcolorbox}

\definecolor{cecol}{HTML}{F1F1F1} 
\definecolor{matrixcol}{HTML}{EDF2F6}  
\definecolor{regucol}{HTML}{e9e5e6}  
\definecolor{samcol}{HTML}{dae4d9}   
\definecolor{semicol}{HTML}{D8E6EA}
\def\tsc#1{\csdef{#1}{\textsc{\lowercase{#1}}\xspace}}
\tsc{WGM}
\tsc{QE}
\tsc{EP}
\tsc{PMS}
\tsc{BEC}
\tsc{DE}

\begin{document}
\let\WriteBookmarks\relax
\def\floatpagepagefraction{1}
\def\textpagefraction{.001}

\setcounter{topnumber}{5}
\setcounter{bottomnumber}{5}
\setcounter{totalnumber}{10}
\setcounter{dbltopnumber}{3}
\shorttitle{LNMBench}
\shortauthors{Yuan Ma et~al.}

\title [mode = title]{Benchmarking Real-World Medical Image Classification with Noisy Labels: Challenges, Practice, and Outlook}                      



\author[1]{Yuan Ma}
\fnmark[1]

\ead{s2420410@jaist.ac.jp}


\affiliation[1]{
  organization={Japan Advanced Institute of Science and Technology},
  city={Nomi},
  country={Japan}
}
\affiliation[2]{
  organization={University College London},
  city={London},
  country={United Kingdom}
}
\affiliation[3]{
  organization={The Hong Kong University of Science and Technology},
  city={Hong Kong},
  country={China}
}
\affiliation[4]{
  organization={University of Toyama},
  city={Toyama},
  country={Japan}
}
\affiliation[5]{
  organization={Monash University},
  city={Melbourne},
  country={Australia}
}
\author[3]{Junlin Hou}
\fnmark[1]
\ead{csejlhou@ust.hk}

\author[4]{Chao Zhang}
\ead{zhang@eng.u-toyama.ac.jp}
\author[2]{Yukun Zhou}
\ead{yukun.zhou.19@ucl.ac.uk}

\author[5]{Zongyuan Ge}
\ead{Zongyuan.Ge@monash.edu}
\author[1]{Haoran Xie}
\cormark[1]
\ead{xie@jaist.ac.jp}
\author[2,5]{Lie Ju}
\ead{lie.ju@ucl.ac.uk}
\cormark[1]
\cortext[cor1]{Corresponding author}
\fntext[1]{These authors contributed equally to this work.}

\begin{abstract}
Learning from noisy labels remains a major challenge in medical image analysis, where annotation demands expert knowledge and substantial inter-observer variability often leads to inconsistent or erroneous labels. 
Despite extensive research on learning with noisy labels (LNL), the robustness of existing methods in medical imaging has not been systematically assessed. 
To address this gap, we introduce LNMBench, a comprehensive benchmark for \textbf{L}abel \textbf{N}oise in \textbf{M}edical imaging. 
LNMBench encompasses \textbf{10} representative methods evaluated across \textbf{7} datasets, \textbf{6} imaging modalities, and \textbf{3} noise patterns, establishing a unified and reproducible framework for robustness evaluation under realistic conditions.
Comprehensive experiments reveal that the performance of existing LNL methods degrades substantially under high and real-world noise, highlighting the persistent challenges of class imbalance and domain variability in medical data. 
Motivated by these findings, we further propose a simple yet effective improvement to enhance model robustness under such conditions. 
The LNMBench codebase is publicly released to facilitate standardized evaluation, promote reproducible research, and provide practical insights for developing noise-resilient algorithms in both research and real-world medical applications.
The codebase is publicly available on https://github.com/myyy777/LNMBench.
\end{abstract}




\maketitle
\begin{figure*}[t]
  \centering
  \begin{subfigure}[t]{\textwidth}
    \centering
    \includegraphics[width=\textwidth]{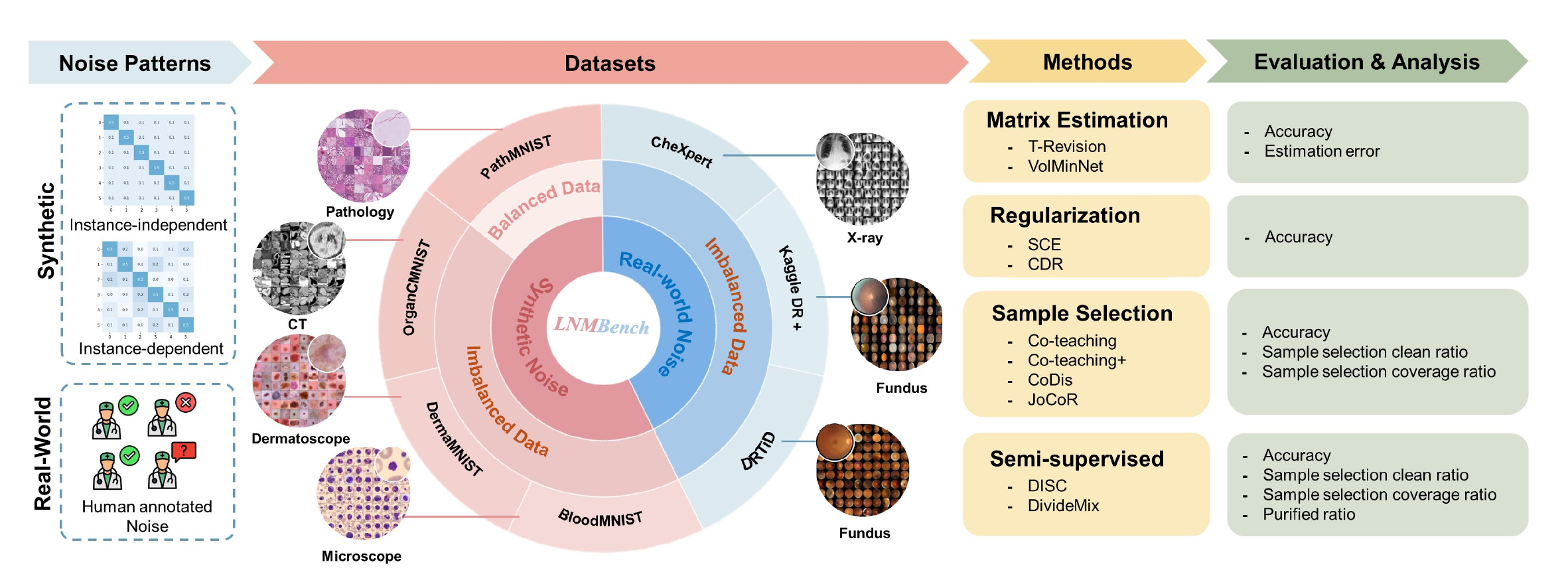}
    \caption{Overview of LNMBench.}
    \label{fig:overview}
  \end{subfigure}
  \vspace{0.3em}  

  \begin{subfigure}[t]{0.24\textwidth}
    \centering
    \includegraphics[width=\linewidth]{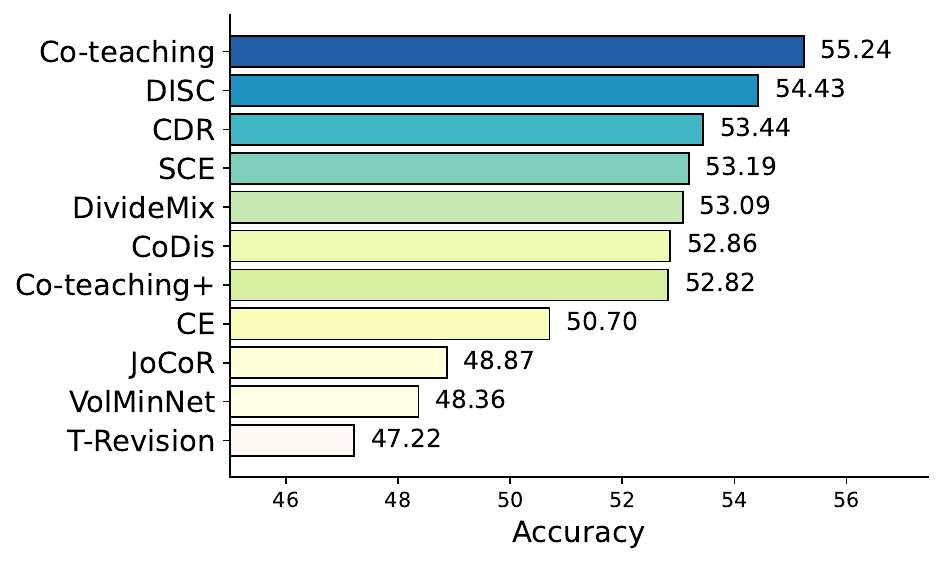}
    \caption{Overall rank of LNMBench.}
  \end{subfigure}
  \begin{subfigure}[t]{0.24\textwidth}
    \centering
    \includegraphics[width=\linewidth]{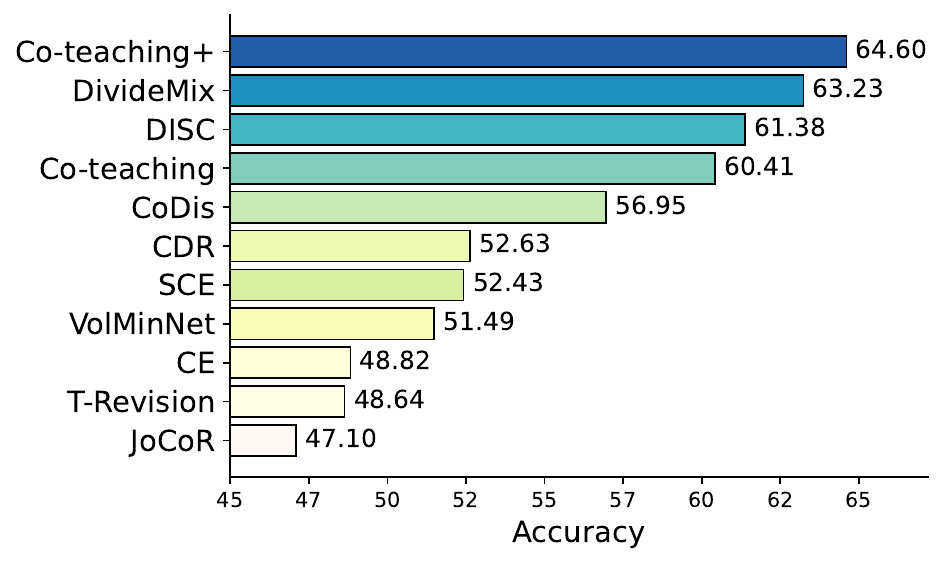}
    \caption{Symmetric noise.}
  \end{subfigure}
  \begin{subfigure}[t]{0.24\textwidth}
    \centering
    \includegraphics[width=\linewidth]{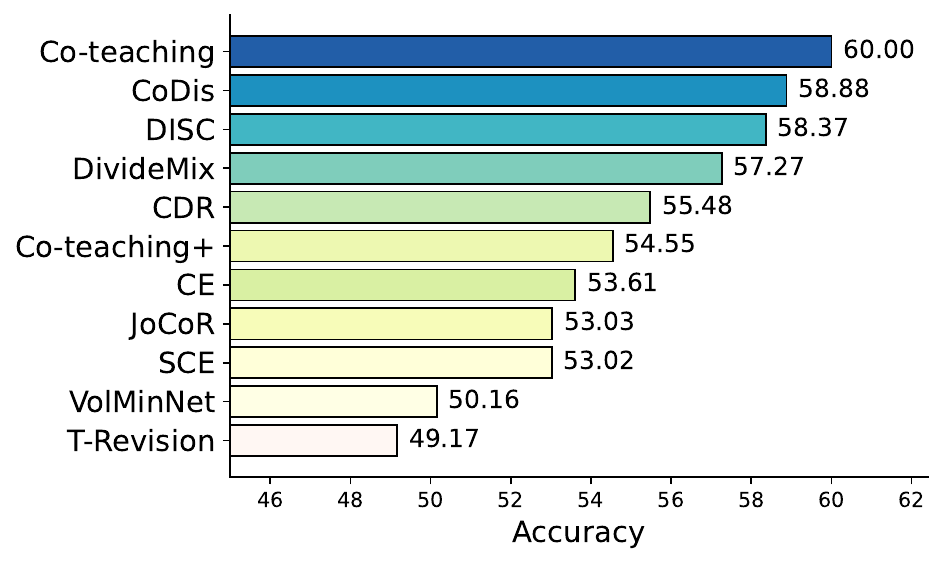}
    \caption{Instance-dependent noise.}
  \end{subfigure}
  \begin{subfigure}[t]{0.24\textwidth}
    \centering
    \includegraphics[width=\linewidth]{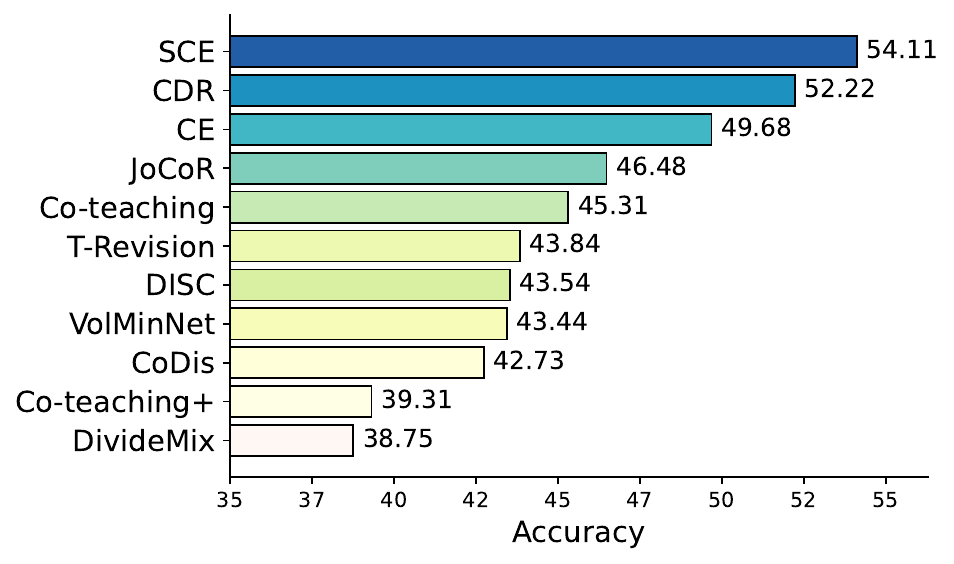}
    \caption{Real-world noise.}
  \end{subfigure}

  \caption{(a)~We present a systematic benchmark for learning with noisy labels in medical image analysis, encompassing 3 noise patterns, 7 datasets from 6 imaging modalities, and 10 representative methods. Integrating both synthetic and real-world noise across balanced and imbalanced datasets, the benchmark provides a unified and reproducible framework that evaluates not only classification accuracy but also robustness-oriented metrics such as estimation error for matrix estimation methods. (b)~Overall rank of LNMBench. (c)~Overall results of LNL methods under symmetric noise. (d)~Overall results of LNL methods under instance-dependent noise. (e)~Overall results of LNL methods under real-world noise.}
  \label{fig:teaser}
\end{figure*}

\input{intro}
\input{related}
\input{preliminaries}
\input{experiments}
\input{conclusion}
\section*{Acknowledgements}
This work is supported by JST SPRING JPMJSP2102 and JST BOOST JPMJBY24D6.
\bibliographystyle{cas-model2-names}

\bibliography{cas-refs}





\end{document}

%% file: intro.tex
\section{Introduction}
\label{sec:introduction}
Deep learning has become the dominant paradigm in computer vision because of its ability to learn expressive feature representations from large-scale annotated datasets.
The performance of deep neural networks (DNNs) is heavily dependent on the quality and quantity of annotated data. 
As a result, robust learning under label noise has attracted considerable attention.
In recent years, a wide range of LNL methods have been developed~\cite{sukhbaatar2014training,tanno2019learning,chen2019understanding,nguyen2019self}. 
Several works aim to model the label corruption process explicitly, typically by estimating a noise transition matrix that reflects the probability of label flips~\cite{xia2019anchor,liao2025instance}. 
Robust loss functions and regularization methods have been proposed to reduce the impact of incorrect supervision during optimization~\cite{wang2019symmetric,labelsmoothing}.
A common strategy for LNL is to select potentially clean samples for training, often based on the small-loss criterion, which assumes that clean examples tend to yield lower loss in early training stages~\cite{han2018co,yu2019does}. 
In addition, semi-supervised learning methods~\cite{li2023disc,li2020dividemix} further build upon sample selection by leveraging the selected clean data as labeled examples and treating the remaining uncertain data as unlabeled, thereby improving robustness through consistency regularization and pseudo labeling. 
These methods have achieved promising results on natural images but lack a systematic evaluation on medical images.

In medical imaging, label noise is particularly prevalent, as annotation requires specialized expertise, and variations in doctors’ experience as well as inter-annotator variability may result in inconsistent or even incorrect labels.
Therefore, learning from noisy labels has long been recognized as an important problem  in medical imaging domain~\cite{xue2019robust,hou2025qmix,pham2021interpreting,ghesu2019quantifying,shi2024survey}.
Early studies attempted to address the label noise problem in medical image analysis through noise-robust regularization methods~\cite{dgani2018training,xue2019robust}.
For instance, Dgain et al.~\cite{dgani2018training} proposed a neural network training strategy that explicitly models label noise within the architecture, and demonstrated its effectiveness on breast microcalcification classification from multi-view mammograms, outperforming standard training methods.
In a comprehensive survey, Karimi et al.~\cite{karimi2020deep} discussed the challenges posed by label noise in deep learning for medical imaging, emphasizing its prevalence, negative impact on performance, and possible mitigation strategies.
Recently, LNL methods have focused on mitigating the effects of label noise by selecting or weighting samples that are likely to be correctly labeled~\cite{ju2022improving,hou2025qmix,khanal2023improving}.
Specifically, Ju et al.~\cite{ju2022improving} proposed a dual-uncertainty estimation and curriculum boosting framework to address the inter-observer disagreement inherent in medical image annotation.
Most of these studies are designed for specific datasets, use different evaluation metrics, and are not released for broader use.
Therefore, conducting fair, systematic, and publicly accessible comparisons is essential for evaluating and advancing LNL methods in medical imaging.

To address this gap, we establish LNMBench, a comprehensive benchmark that systematically evaluates 10 representative LNL methods on 7 medical imaging datasets (see Fig.~\ref{fig:teaser}).
LNMBench is organized around the progression from noise patterns to datasets, methods, and evaluation, aiming to provide a holistic understanding of how existing algorithms perform under various noise conditions in medical imaging.
We categorize label noise into two major sources: synthetic and real-world.
To evaluate performance across diverse imaging modalities and class distributions, we adopt seven medical datasets, including pathology slides, dermatology images, fundus photography, microscope images, CT and chest X-rays.
LNMBench covers both balanced datasets~(PathMNIST~\cite{medmnistv2}), which provide reference performance under approximately symmetric class distributions, and imbalanced datasets~(DermaMNIST, BloodMNIST, and OrganCMNIST~\cite{medmnistv2}), which reflect the inherent class imbalance characteristic of medical imaging~\cite{ju2023hierarchical}.
We inject synthetic noise at different ratios to both groups.
To assess model robustness under real clinical conditions, we introduce real-world noisy datasets, DRTiD~\cite{hou2022cross}, Kaggle DR+~\cite{ju2022improving}, and CheXpert~\cite{johnson2019mimic} that contain clinically sourced label noise.
LNMBench covers widely used LNL methods and one specifically designed for class imbalanced data.
For evaluation, we report classification accuracy as well as robustness-oriented metrics, such as the clean ratio and coverage ratio for sample selection–based methods.
Experimental results demonstrate that current LNL methods are inadequate for effectively addressing label noise in medical imaging, particularly in high and real-world noise~(see Fig.~\ref{fig:teaser}).
In addition, we reveal the impact of class imbalance on the performance of sample selection based methods.
Based on extensive experimental results, we rank the methods in LNMBench by their classification accuracy across diverse noise patterns and provide an overall performance comparison (see Fig.~\ref{fig:teaser}).
Building on these insights, we summarize the limitations of existing methods and outline potential directions for future research.
To address real-world challenges in medical image analysis, we propose MedSSL, a general strategy applicable to semi-supervised methods.
Specifically, MedSSL incorporates label smoothing in warm-up stage, introduces a regularization term during training, and employs class-specific thresholds for adaptive sample selection.
Experimental results demonstrate that MedSSL improves the performance of semi-supervised methods on real-world noisy datasets.

Our main contributions from this work are summarized as follows:
\begin{itemize}
    \item We introduced a systematic benchmark for LNL methods in medical image analysis, encompassing 3 noise patterns, 7 medical datasets spanning from 6 medical domains, and 10 representative methods. This benchmark provided a standardized foundation for evaluating robustness under diverse conditions and offers valuable insights for future research.
    \item We ranked the methods in LNMBench by their classification accuracy across various noise patterns.
    Furthermore, we provided a comprehensive analysis of the strengths and weaknesses of existing methods, and outline potential directions for future research. 
    Our findings and insights will inspire the development of more effective LNL methods for medical data.
    \item We developed a well-structured and user-friendly codebase for LNL, which supports 10 LNL methods and allows flexible substitution of backbone networks and evaluation metrics.
\end{itemize} 

%% file: related.tex
\section{Related Work}
\label{sec:related work}
\subsection{Learning with Noisy Labels}
Learning with noisy labels has received extensive attention in recent years. 
Existing methods can be broadly categorized into noise transition matrix estimation, noise-robust regularization, sample selection, and semi-supervised methods.

\textbf{Noise transition matrix estimation} methods aim to infer clean label distributions from noisy observations and reduce the adverse effects of label corruption.
Early method~\cite{sukhbaatar2014training} modeled label noise by introducing a learnable noise adaptation layer, typically implemented as a linear transformation placed above the softmax output to approximate the label transition process. 
T-Revision~\cite{xia2019anchor} proposed a transition matrix estimation strategy without relying on anchor points. 
However, due to the lack of reliable anchor points in real-world datasets, these methods failed to deliver satisfactory performance.
Recently, researchers have attempted to estimate instance-dependent transition matrices~\cite{liao2025instance,lin2024learning}, allowing the noise modeling to vary across samples which improves robustness in more realistic noisy scenarios.

\textbf{Noise-robust regularization} methods~\cite{wang2019symmetric,hendrycks2019using} aim to enhance noise robustness by constraining the model’s capacity to memorize incorrect labels.
Many works modify loss function for robustness to noisy labels.
SCE~\cite{wang2019symmetric} introduces a combination of Cross Entropy and Reverse Cross Entropy to address both overfitting to noisy labels and underfitting on hard classes, enhancing robustness in noisy label scenarios.
In addition to designing robust losses, recent studies also explore parameter-level strategies to mitigate the impact of noisy labels.
CDR~\cite{xia2020robust} further mitigates noise memorization by identifying critical parameters responsible for fitting clean data and applying targeted regularization, while suppressing updates to non-critical parameters associated with noisy labels.
Beyond loss and parameter-level strategies, some methods improve robustness by directly regularizing the labels~\cite{ko2023gift,labelsmoothing}.
Label smoothing~\cite{labelsmoothing} explores whether the regularization skill of label smoothing is still practical in the presence of label noise.
These methods are highly flexible and easily integrated with other noise-robust techniques, but they often fall short under severe noise conditions.

\textbf{Sample selection} methods aim to find out those samples whose labels are likely to be corrupted~\cite{wei2020combating,xia2023combating}.
Most sample selection strategies conducted noisy label processing by selecting clean samples through a "small-loss" strategy. 
Co-teaching~\cite{han2018co} is a representative sample selection method that trains two networks simultaneously and lets them exchange small-loss instances in each mini-batch to update their own parameters.
Co-teaching+~\cite{yu2019does} extends Co-teaching by introducing a disagreement strategy, where two networks update only on samples with divergent predictions, thus enhancing robustness to high noise by mitigating network consensus on corrupted labels.
These sample selection methods effectively reduce the negative impact of noisy labels by identifying and training on likely clean samples, thereby enhancing model robustness. 
However, they typically rely on prior knowledge of the noise rate, which limits their applicability in practice.

\textbf{Semi-supervised} methods build upon sample selection approaches by treating the identified noisy samples as unlabeled data, and leveraging model predictions as pseudo-labels for subsequent loss calculation~\cite{nguyen2019self,zhou2020robust,chen2023two}.
DivideMix~\cite{li2020dividemix} uses a Gaussian Mixture Model (GMM) to divide samples into clean and noisy sets based on loss, and trains the model with a mix of supervised and semi-supervised learning to reduce the impact of label noise.
DISC~\cite{li2023disc} introduces a dynamic instance-specific strategy for learning with noisy labels by leveraging two-view confidence estimation and per-instance memorization strength. 
Semi-supervised learning-based methods fully exploit the training dataset.
However, due to their reliance on model confidence or memorization heuristics, these methods still struggle to correctly handle hard samples.
\subsection{Medical Image Analysis with Noisy Labels}
In medical image analysis, several attempts have been made to tackle the challenge of learning from label noise~\cite{liao2025unleashing,gehlot2021cnn}.
For instance, Dgani et al.~\cite{dgani2018training} modeled label noise within neural networks to improve mammogram classification.
Similarly, Xue et al.~\cite{xue2019robust} proposed an iterative learning framework with uncertainty based sample mining and re-weighting to handle noisy labels in skin lesion classification.
Pham et al.~\cite{pham2021interpreting} proposed a CNN-based framework with label smoothing to improve thoracic disease classification on the CheXpert dataset~\cite{johnson2019mimic}.
Ghesu et al.~\cite{ghesu2019quantifying} proposed uncertainty-driven bootstrapping to filter training samples with the highest predictive uncertainty, thereby improving robustness and accuracy on the ChestX-Ray8 dataset~\cite{wang2017chestx}.
Ju et al.~\cite{ju2022improving} proposed a dual-uncertainty estimation and curriculum boosting framework to address disagreement and single target label noise in medical image classification.
Some researchers~\cite{li2023learning} have also investigated label noise in imbalanced medical datasets and proposed multi-stage noise removal frameworks to jointly address label noise, class imbalance, and class hardness.
Hou et al.~\cite{hou2025qmix} proposed QMix, a framework for robust disease diagnosis under mixed noise by jointly separating mislabeled and low-quality samples and applying quality-aware semi-supervised training, achieving state-of-the-art results on retinal datasets.
Most of these studies are designed for specific datasets and are not released for broader use.
However, the lack of publicly available and standardized implementations makes it difficult to compare methods fairly, highlighting the importance of establishing a unified benchmark.
\subsection{Benchmarks for Medical Image}
As most existing approaches are designed and tested on natural image datasets, their effectiveness on medical images remains unclear. 
This motivates the need for a dedicated benchmark in the medical imaging domain to ensure proper evaluation and reliable comparisons.
Therefore, many researchers have attempted to establish fair benchmarks for medical images in various fields~\cite{litjens2017survey,shin2016deep,gutbrod2025openmibood}.
Menze et al.~\cite{menze2014multimodal} introduced the BRATS benchmark, a standardized dataset and evaluation framework for brain tumor segmentation using multi-contrast MRI scans.
MedMNIST v2~\cite{medmnistv2} is a large-scale and lightweight benchmark dataset that covers a diverse set of 2D and 3D biomedical images. 
It is designed to enable standardized evaluation across various medical imaging modalities and classification tasks.
MONICA~\cite{ju2024monica} is a comprehensive and unified benchmark for long-tailed learning in medical image analysis. 
It implements over 30 methods and evaluates them on 12 datasets across 6 medical domains, addressing the lack of standardized benchmarks in this area.

Recent studies have attempted to construct benchmarks for investigating label noise in medical imaging~\cite{mehrtens2023benchmarking}.
Karmi et al.~\cite{karimi2020deep} conducted a comprehensive review on label noise in deep learning for medical image analysis, highlighting its overlooked impact and providing experimental insights and practical recommendations for mitigating different types of label noise in medical datasets.
Ju et al. ~\cite{ju2022improving} were the first to propose a unified evaluation framework with consistent hyperparameter settings for systematically comparing different label noise learning methods in the medical image domain.
Several studies have made commendable efforts to establish benchmarks for learning with label noise in the medical imaging domain. 
However, they are evaluated only on specific datasets, cover a limited range of representative methods, and lack systematic analyses of existing challenges and potential research directions.
Therefore, we propose a standardized benchmark to systematically evaluate label noise learning methods under diverse medical imaging scenarios and identify key challenges for future research.

%% file: preliminaries.tex
\section{Definition and Preliminaries}
The goal of this paper is to systematically investigate the effectiveness of representative LNL methods in handling annotation noise within medical image classification tasks. 
In this section, we first introduce the definition of the label noise problem.
We then describe the noise patterns used in our study.  
Finally, we provide a brief introduction to the evaluation metrics.
\subsection{Definition of Label Noise}
We focus on the problem of label noise in image classification. 
Label noise refers to errors in the annotated labels of training samples, leading to discrepancies between the observed label and the underlying true class. 
In a $k$-class classification problem, we are provided with a training dataset
$D = \{(x_i, y_i)\}_{i=1}^N,$
where $x_i \in X$ is the $i$-th image in the feature space $X$, and $y_i \in Y$ is its corresponding observed label in the label space $Y$. 
Due to noise in the annotation process, the observed label $y_i$ may differ from the true label $y_i^* \in Y^*$.  
We denote the probability distribution of class $y$ for an input $x$ as $p(y|x)$, satisfying $\sum_{i=1}^k p(y_i|x) = 1$. 
The goal of a classification task is to learn a mapping function $f: X \to Y$ that minimizes the empirical risk $R$~\cite{cordeiro2020survey}.
\begin{equation}
R = \mathbb{E}_{(x,y)\in D}\big[L(f(x), y)\big] 
  = \mathbb{E}_{x,y_x}\big[L(f(x), y_x)\big],
\end{equation}
where $L$ is the loss function and the expectation $\mathbb{E}$ is taken over the data distribution.  
However, in the presence of label noise, the empirical risk is no longer noise-tolerant, as the computation of the loss involves incorrect labels. 
As a result, deep neural networks may overfit these noisy labels, leading to corrupted feature representations and degraded generalization performance.  
\subsection{Noise Patterns}
\textbf{Instance-independent label noise} refers to a noise pattern where the label corruption process does not depend on the input instance features, but only on the original class label. 
Two representative types of this noise are symmetric and asymmetric label noise. 
We model label noise using a transition matrix $\tau \in \mathbb{R}^{k \times k}$, where $\tau_{ij}$ denotes the probability that a sample with true label $i$ is observed as label $j$, and $k$ represents the total number of classes. Let $\mathcal{A}$ denote the set of all class indices, $\eta$ denote the noise rate.
Symmetric noise is modeled by randomly choosing a label from all classes 
including the ground truth label, which means 
$\forall j \in \mathcal{A}, \;\tau_{ij} = \tfrac{\eta}{k - 1}$. 
In asymmetric noise, the ground truth label is not included in the label 
flipping options, where $i \neq j$, 
$\tau_{ij} = \tfrac{\eta}{k - 1}$. 
In contrast to symmetric noise, asymmetric noise represents a noise process where the ground truth label is flipped into one  specific label whose class is closer to the true label.
We adopt symmetric noise in our benchmark because it provides a controlled and widely used corruption pattern, enabling systematic evaluation and fair comparison across different methods.

\textbf{Instance-dependent label noise} refers to a more realistic type of label corruption where the noise generation process depends on the input data features. 
Even among instances belonging to the same class, significant variations in visual can lead to different probabilities of being mislabeled. 
As a result, the label flipping probability is not uniform across instances, but rather highly conditioned on each specific sample. 
This property makes instance dependent noise better suited for modeling real world annotation errors.
A commonly used approach to generate instance-dependent label noise is to assign each sample a flipping probability based on its input features, and then compute a probability distribution over classes by projecting the features onto randomly sampled vectors~\cite{xia2020part}. 
A noisy label is then sampled from this distribution. 
This method allows different instances from the same class to have different flipping behaviors.
However, the direction vectors are sampled randomly, so the synthetic noise may not be suitable for simulating the types of errors made by human experts. 



To address this limitation, we introduce a new instance-dependent label noise generation strategy that leverages model predictions and uncertainty to construct more realistic label noise (see Algorithm~\ref{idn}).
Let $f$ be a network trained on a clean reference set, yielding soft predictions $p(j\mid x)$. 
For each training sample $(x_i,y_i)$, we mask the ground-truth class and renormalize the remaining probabilities to obtain a mislabeling distribution $\pi(\cdot\mid x_i)$. 
Given a sample-specific flip rate $q_i\in[0,1]$ (with mean $\eta$), the per-instance transition is
\[
T_i(x_i)=(1-q_i)\,\mathbf e_{y_i}+q_i\,\pi(x_i),
\]
where $\mathbf e_{y_i}$ is the one-hot vector corresponding to the true class $y_i$.
The noisy label is then drawn as $\tilde y_i\sim\mathrm{Categorical}\big(T_i(x_i)\big)$.
This formulation produces feature-conditioned, heterogeneous flips that better reflect realistic annotation errors.
\begin{algorithm}[ht]
\caption{Instance-dependent Label Noise Generation}
\label{idn}

\textbf{Input:} Test set $D_{\text{test}}$; Train set $D_{\text{train}} = \{(x_i, y_i)\}_{i=1}^N$; noise rate $\eta$. \\

\textbf{Step 1: Train a clean model} \\
Train a classifer $f$ on clean test set $D_{\text{test}}$; \\

\textbf{Step 2: Generate instance-dependent noisy labels} \\
Sample instance flip rates $q$ from truncated normal distribution $\mathcal{N}(\eta, 0.1^2, [0, 1])$; \\
\textbf{For} $i = 1, 2, \dots, N$ \textbf{do} \\
\hspace{1em} $z = f(x_i)$  \\
\hspace{1em} $z_{y_i} = -\infty$  \\
\hspace{1em} $p = q_i \times \mathrm{softmax}(z)$  \\
\hspace{1em} $\tilde{p}_{y_i} = 1 - q_i$   \\
\hspace{1em} Sample a noisy label $y_i$ from the label space \\
\hspace{1em} according to probabilities $p$ \\
\textbf{End for} \\
\textbf{Output:} Noisy samples $\tilde{D} = \{(x_i, \tilde{y}_i)\}_{i=1}^N$ \\
\end{algorithm}

\textbf{Real-world label noise} refers to the annotation errors that naturally arise when datasets are labeled by human annotators.
Previous synthetic label noise models, while useful for controlled experiments, fall short in capturing the complexity of real-world annotation errors~\cite{wei2021learning}. 
Medical datasets often involve ambiguous cases and limited expert consensus, making label noise more subtle and difficult to replicate through simple simulations.
Therefore, evaluating label noise learning methods on real-world medical datasets where annotation uncertainty reflects actual clinical practice is more meaningful for assessing their true effectiveness.

\subsection{Robust Metrics under LNL}

Checkpoint selection under label noise is a challenging problem because models often learn meaningful patterns in early epochs but eventually overfit noisy labels. 
Most existing studies~\cite{ju2022improving,hou2025qmix} evaluate performance using either the best test accuracy (B), which reflects the upper bound of model capacity, or the average test accuracy over the last few epochs (L), which captures training stability. 
However, both require access to the test set and therefore do not reflect realistic model selection.
To approximate practical conditions, we further consider validation-selected accuracy (V), defined as the test accuracy at the epoch where the noisy validation accuracy is maximized. Reporting V allows us to assess whether checkpoint selection based on a noisy validation set is reliable. 
By comparing V against B and L, we quantify the gap between practical selection and ideal or stable performance.
In this work, we report all three metrics (B, L, V) to ensure comparability with prior studies and to provide direct evidence of the reliability of noisy validation-based selection. 
Each experiment is repeated with three random seeds, and averaged results are reported for robustness.

%% file: experiments.tex
\begin{table*}
\caption{Performance comparison under different noise types on PathMNIST. \textbf{Bold} indicates the best result}
\label{pathmnist_results}
\small

\centering
\begin{tblr}{
  cell{1}{3} = {c=6}{},
  cell{4}{1} = {r=2}{},
  cell{6}{1} = {r=3}{},
  cell{9}{1} = {r=3}{},
  cell{12}{1} = {r=3}{},
  cell{15}{1} = {r=3}{},
  cell{18}{1} = {r=3}{},
  cell{21}{1} = {r=3}{},
  cell{24}{1} = {r=3}{},
  cell{27}{1} = {r=3}{},
  colspec={c c *{6}{c}},
  rows = {abovesep=0.2ex, belowsep=0.2ex},  
  colsep = 3pt,                              
  row{1} = {font=\bfseries, halign=c},
  hline{1,2,3,Z} = {-}{},  
  row{3-5} = {bg=cecol},
row{6-11} = {bg=matrixcol},
row{12-17} = {bg=regucol},
row{18-29} = {bg=samcol},
row{30-35} = {bg=semicol},
}
Dataset    &        & \SetCell[c=6]{c} PathMNIST \\
Noise type &        & Sym-20\% & Sym-50\% & Sym-90\% & Idn-20\% & Idn-50\% & Idn-90\% \\
        & B & 94.40 ± 0.37 & 93.82 ± 0.91 & 13.42 ± 1.82 & 93.34 ± 1.24 & 76.63 ± 4.24 &  20.77 ± 1.88 \\
  CE       & V & 91.34 ± 3.34 & 92.99 ± 1.66  & 9.69 ± 2.05  & 93.39 ± 1.24 & 76.34 ± 4.38 & 11.14 ± 2.79 \\
           & L & 84.01 ± 0.85 & 52.25 ± 1.93  & 10.21 ± 0.93 & 86.00 ± 0.61 & 56.54 ± 0.36  & 15.13 ± 1.10 \\
\midrule
T-Revision   & B    & 96.13 $\pm$ 0.91    & 91.23 $\pm$ 0.35    & 8.25 $\pm$ 1.63     & 91.63 $\pm$ 0.33    & 74.09 $\pm$ 0.72    & 23.50 $\pm$ 1.68 \\
             & V    & 94.87 $\pm$ 0.84    & 87.23 $\pm$ 0.77    & 8.25 $\pm$ 1.13     & 91.51 $\pm$ 0.56    &  63.40 $\pm$ 0.90   & 11.34 $\pm$ 1.53 \\
             & L    & 85.30 $\pm$ 0.79    & 52.37 $\pm$ 0.65    & 8.25 $\pm$ 1.87     & 87.06 $\pm$ 0.91    & 53.13 $\pm$ 0.44    & 11.17 $\pm$ 0.96 \\
\midrule

 VolMinNet              & B    & \textbf{96.16 $\pm$ 0.49} & 92.97 $\pm$ 0.37    & 15.81 $\pm$ 1.53    & 91.53 $\pm$ 0.35    & 77.18 $\pm$ 1.20    & 17.92 $\pm$ 2.55 \\
  & V    & \textbf{96.03 $\pm$ 0.23} & 92.81 $\pm$ 0.68    & 6.59 $\pm$ 0.92     & 90.78 $\pm$ 0.87    & 72.17 $\pm$ 0.38    & 13.50 $\pm$ 1.77 \\
             & L    & 88.01 $\pm$ 0.88    & 56.92 $\pm$ 0.69   & 9.53 $\pm$ 0.99     & 84.63 $\pm$ 0.87   & 54.03 $\pm$ 0.54    & 13.63 $\pm$ 1.10 \\
\midrule
SCE          & B    & 94.94 $\pm$ 0.31    & 94.60 $\pm$ 1.10   & 11.18 $\pm$ 3.28   & 95.10 $\pm$ 0.45    & 90.36 $\pm$ 0.72    & 16.44 $\pm$ 4.23 \\
             & V    & 94.42 $\pm$ 1.52   & 94.31 $\pm$ 0.79    & 8.21 $\pm$ 2.31    & \textbf{95.10 $\pm$ 0.89} & \textbf{90.35 $\pm$ 0.85 }  & 7.89 $\pm$ 2.17 \\
             & L    & 90.94 $\pm$ 0.71    & 58.84 $\pm$ 1.38    & 7.50 $\pm$ 2.55     & 87.37 $\pm$ 2.54    & 56.23 $\pm$ 2.75    & 16.23 $\pm$ 2.43 \\
\midrule
CDR          & B    & 94.79 $\pm$ 0.47   & 95.15 $\pm$ 0.86  & 11.47 $\pm$ 1.94    & 95.07 $\pm$ 0.54    & 81.66 $\pm$ 0.25    & 20.85 $\pm$ 1.21 \\
             & V    & 91.64 $\pm$ 0.29    & \textbf{94.97} $\pm$ 0.81 & 11.47 $\pm$ 1.14    & 94.81 $\pm$ 0.24    & 81.66 $\pm$ 0.35    & 7.66 $\pm$ 1.88 \\
             & L    & 89.73 $\pm$ 0.14    & 57.94 $\pm$ 2.11   & 10.23 $\pm$ 1.39    & 89.25 $\pm$ 0.28    & 59.10 $\pm$ 1.72    & 15.84 $\pm$ 2.01 \\
\midrule
Co-teaching           & B & 93.76 $\pm$ 0.66 & 92.78 $\pm$ 0.09 & 16.57 $\pm$ 2.37 & 94.14 $\pm$ 0.46 & 91.93 $\pm$ 0.97 & 25.20 $\pm$ 0.87 \\
& V & 89.55 $\pm$ 1.56 & 87.67 $\pm$ 5.00 &  5.52 $\pm$ 3.53 & 90.34 $\pm$ 3.30 & 89.63 $\pm$ 1.33 & \textbf{17.34 $\pm$ 1.79} \\
           & L & 89.63 $\pm$ 0.78 & 82.79 $\pm$ 0.57 &  8.82 $\pm$ 3.28 & 90.49 $\pm$ 0.77 & \textbf{86.54 $\pm$ 0.97} & 15.50 $\pm$ 5.89 \\

\midrule

Co-teaching+ & B    & 93.50 $\pm$ 0.39 & 93.39 $\pm$ 1.37 & 16.21 $\pm$ 1.89 & 94.00 $\pm$ 0.37 & 82.63 $\pm$ 5.82 & \textbf{29.52 $\pm$ 2.15} \\
             & V    & 90.85 $\pm$ 1.78 & 92.26 $\pm$ 1.16 & \textbf{11.71 $\pm$ 1.21 } & 92.21 $\pm$ 1.14 & 81.28 $\pm$ 7.91 & 14.49 $\pm$ 1.41 \\
             & L    & 91.13 $\pm$ 1.34 & \textbf{91.87 $\pm$ 0.98} & 10.81 $\pm$ 0.85 & \textbf{92.30 $\pm$ 0.27} & 66.09 $\pm$ 8.35 & \textbf{20.17 $\pm$ 1.17} \\
\midrule

CoDis & B    & 94.66 $\pm$ 0.57 & 94.40 $\pm$ 1.94 & \textbf{24.53 $\pm$ 2.48} & 93.51 $\pm$ 0.67 & 91.84 $\pm$ 1.32 & 26.21 $\pm$ 3.28 \\
             & V    & 92.44 $\pm$ 1.13 & 88.30 $\pm$ 1.11 & 2.26 $\pm$ 1.49  & 93.48 $\pm$ 1.26 & 87.72 $\pm$ 1.18 & 15.26 $\pm$ 1.08 \\
             & L    & 89.19 $\pm$ 1.32 & 80.09 $\pm$ 1.76 & 1.58 $\pm$ 2.71 & 87.58 $\pm$ 1.31 & 86.07 $\pm$ 1.73 & 13.61 $\pm$ 2.56 \\
\midrule

JoCoR        & B    & 92.03 $\pm$ 0.61 & 91.17 $\pm$ 1.25 & 16.26 $\pm$ 3.09 & 92.16 $\pm$ 1.60 & 81.33 $\pm$ 3.77 & 24.93 $\pm$ 2.74 \\
             & V    & 88.19 $\pm$ 0.91 & 89.96 $\pm$ 0.84 &  7.22 $\pm$ 5.07 & 90.09 $\pm$ 1.24 & 71.05 $\pm$ 3.63 & 12.55 $\pm$ 1.07 \\
             & L    & 83.10 $\pm$ 0.31 & 49.84 $\pm$ 1.12 &  8.37 $\pm$ 3.07 & 82.57 $\pm$ 1.41 & 54.49 $\pm$ 0.90 & 15.99 $\pm$ 0.89 \\
\midrule
           & B & 94.14 $\pm$ 0.79 & 93.80 $\pm$ 1.27 & 19.09 $\pm$ 2.31 & 93.28 $\pm$ 0.55 & 86.83 $\pm$ 0.68 & 18.98 $\pm$ 4.35 \\
DISC       & V & 88.72 $\pm$ 0.97 & 90.43 $\pm$ 2.01 &  9.16 $\pm$ 1.64 & 88.19 $\pm$ 2.49 & 84.14 $\pm$ 2.61 &  4.69 $\pm$ 0.60 \\
           & L & 87.76 $\pm$ 1.03 & 90.58 $\pm$ 3.01 &  5.80 $\pm$ 6.11 & 87.98 $\pm$ 1.98 & 82.11 $\pm$ 2.76 &  4.76 $\pm$ 0.67 \\
\midrule
    & B    & 95.79 $\pm$ 0.40 & \textbf{95.26 $\pm$ 0.47} & 12.52 $\pm$ 0.10 & \textbf{96.13 $\pm$ 0.65} & \textbf{92.29 $\pm$ 0.11} & 16.02 $\pm$ 1.49 \\
DivideMix          & V    & 93.86 $\pm$ 0.37 & 90.18 $\pm$ 2.23 &  5.37 $\pm$ 4.26 & 92.85 $\pm$ 1.03 & 88.38 $\pm$ 0.56 & 11.52 $\pm$ 1.52 \\
             & L    & \textbf{91.39 $\pm$ 0.33} & 91.29 $\pm$ 0.05 &  7.02 $\pm$ 2.21 & 90.33 $\pm$ 0.27 & 82.35 $\pm$ 2.96 &  7.66 $\pm$ 3.41 \\
\end{tblr}
\end{table*}



{
\captionsetup[subfigure]{font=footnotesize}
\begin{figure*}[t]
    \centering
    \begin{subfigure}[b]{0.13\textwidth}
        \centering
        \includegraphics[width=\textwidth]{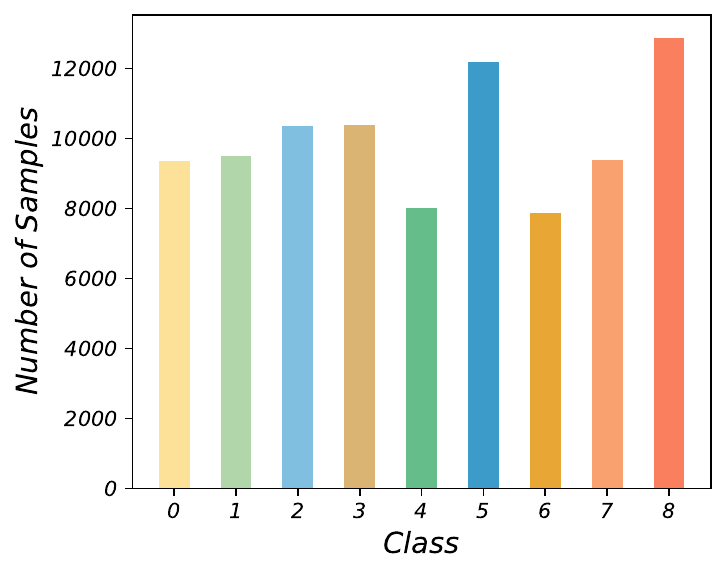}
        \caption{PathMNIST}
    \end{subfigure}
    \begin{subfigure}[b]{0.13\textwidth}
        \centering
        \includegraphics[width=\textwidth]{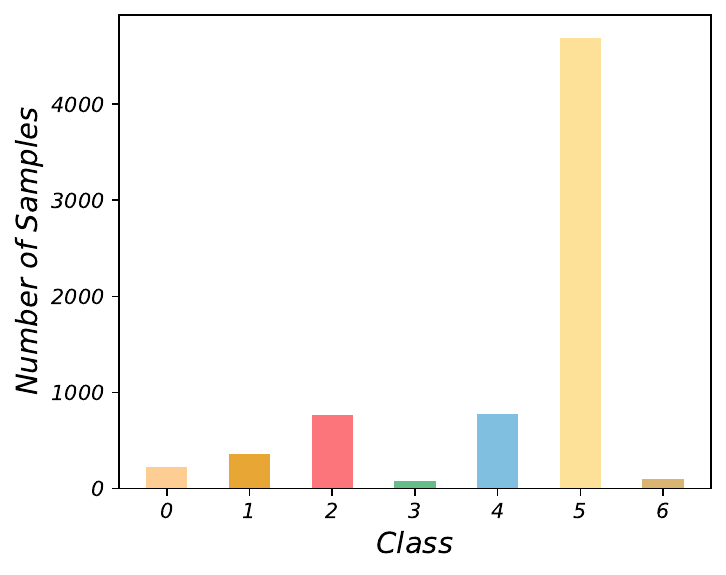}
        \caption{DermaMNIST}
    \end{subfigure}
    \begin{subfigure}[b]{0.13\textwidth}
        \centering
        \includegraphics[width=\textwidth]{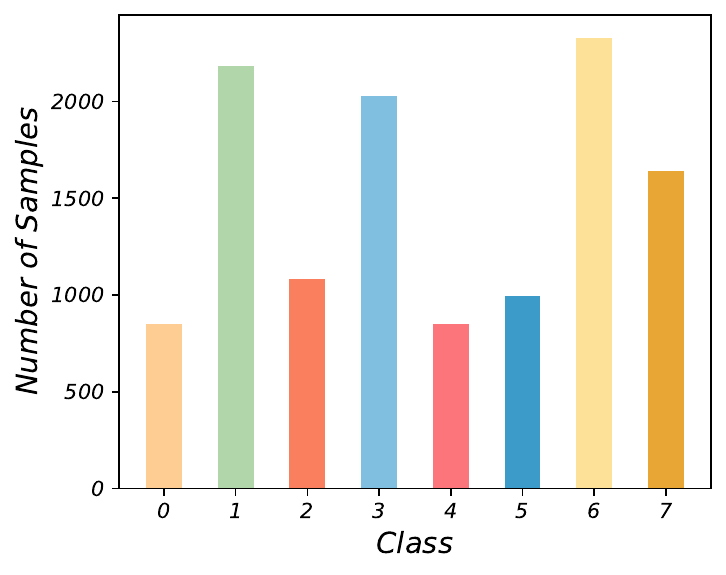}
        \caption{BloodMNIST}
    \end{subfigure}
    \begin{subfigure}[b]{0.13\textwidth}
        \centering
        \includegraphics[width=\textwidth]{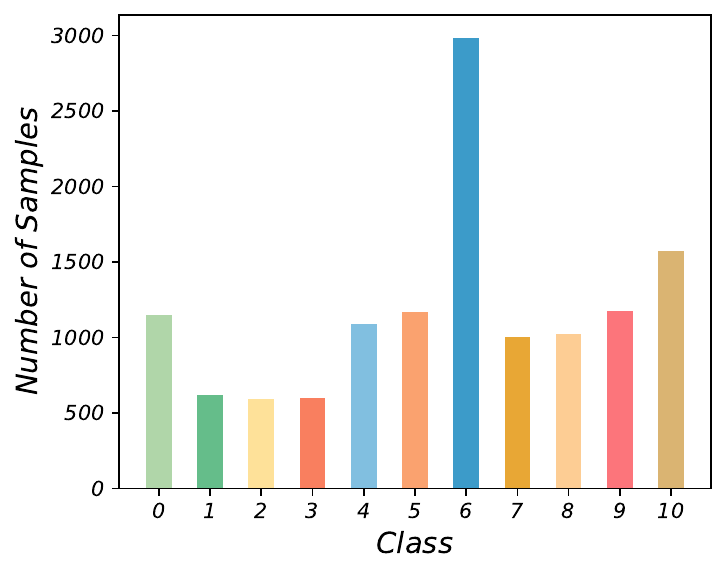}
        \caption{OrganCMNIST}
    \end{subfigure}
    \begin{subfigure}[b]{0.13\textwidth}
        \centering
        \includegraphics[width=\textwidth]{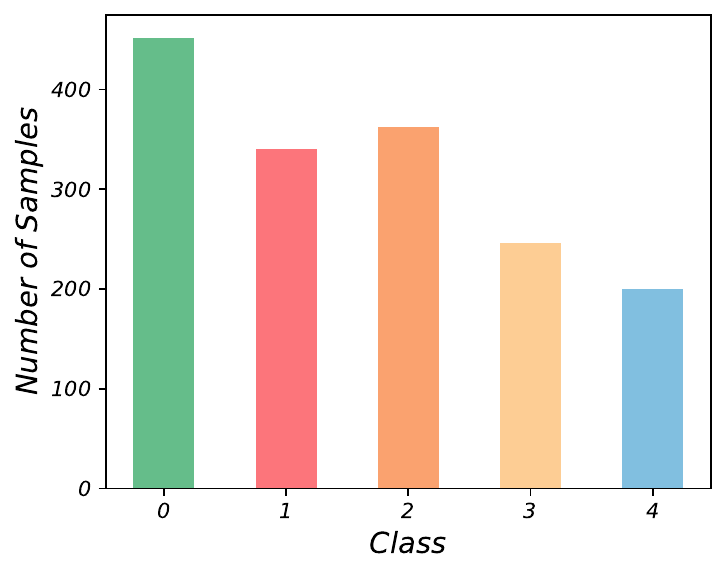}
        \caption{DRTiD}
    \end{subfigure}
    \begin{subfigure}[b]{0.13\textwidth}
        \centering
        \includegraphics[width=\textwidth]{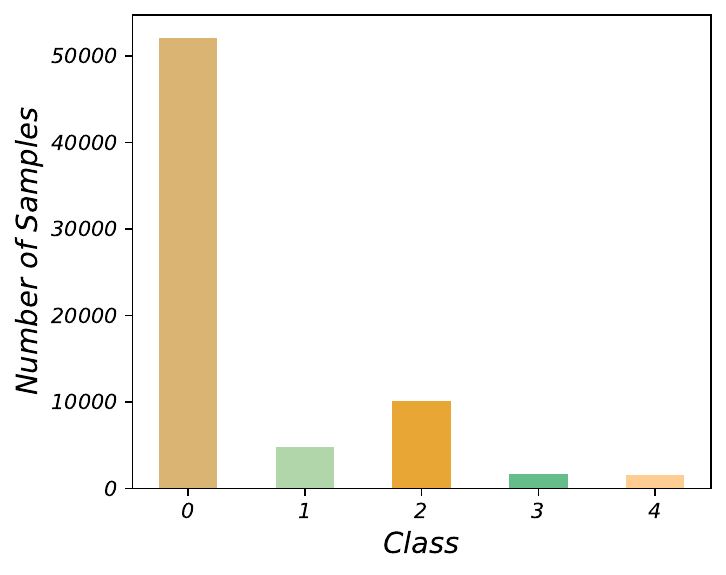}
        \caption{KaggleDR+}
    \end{subfigure}
    \begin{subfigure}[b]{0.13\textwidth}
        \centering
        \includegraphics[width=\textwidth]{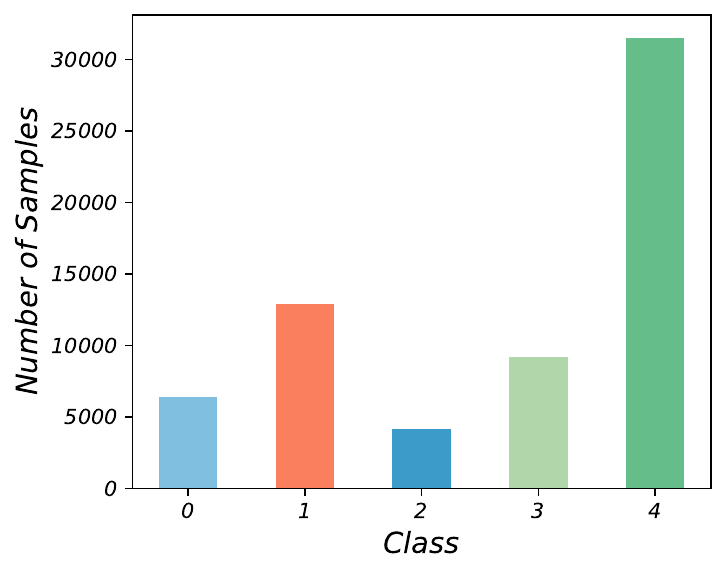}
        \caption{CheXpert}
    \end{subfigure}

    \caption{Overview of class distributions across the datasets utilized in LNMBench.}
    \label{fig:datadistribution}
\end{figure*}
}

\section{Experiments}
We present extensive experiments to evaluate the robustness of representative LNL methods on medical image classification tasks. 
We aim to examine their effectiveness under different noise types, ratios, and dataset characteristics.
In Section~\ref{datasec}, we introduce the datasets used in LNMBench.
Section~\ref{methodsec} provides description of the evaluated methods.
Section~\ref{detailsec} presents the implementation details of our experiments.
In Section~\ref{overallsec}, we summarize the overall experimental results.
Section~\ref{balancesec} reports the results on balanced datasets with three ratios of symmetric and instance-dependent noise.
Section~\ref{imbalancesec} reports the results on imbalanced datasets with three ratios of instance-dependent noise.
Section~\ref{realworldsec} reports the results on real-world imbalanced noisy datasets.
Section~\ref{summarysec} discusses the key challenges identified in current methods and provides insights for future research directions.
\subsection{Datasets}\label{datasec}
As shown in the Fig.~\ref{fig:datadistribution}, we illustrate the class distributions of the datasets utilized in LNMBench.
To systematically assess robustness, we consider three evaluation regimes. First, we use the near-balanced benchmark PathMNIST~\cite{medmnistv2} to establish reference performance under approximately symmetric class distributions. 
Second, given that class imbalance is pervasive in medical imaging, we include DermaMNIST, BloodMNIST, and OrganCMNIST~\cite{medmnistv2} to examine performance under realistic distribution skew. 
For these two groups, we inject synthetic noise (symmetric and instance-dependent). 
Finally, we evaluate external validity under clinically sourced label noise using the real-world noise datasets DRTiD~\cite{hou2022cross}, Kaggle DR+~\cite{ju2022improving}, and CheXpert~\cite{johnson2019mimic}.

\subsubsection{Balanced Dataset}
PathMNIST~\cite{medmnistv2} is derived from colorectal cancer histology slides and consists of hematoxylin and eosin stained tissue patches for a nine-class classification task.
Each sample is a $224\times224$ RGB image cropped from whole-slide scans.
The dataset contains 107,180 images, officially divided into 89,996 training, 10,004 validation, and 7,180 test samples.

\subsubsection{Imbalanced Datasets}
\textbf{DermaMNIST}~\cite{medmnistv2} contains 10,015 dermatoscopic images of common pigmented skin lesions (size $224\times224$), categorized into seven classes.
The official split includes 7,008 training, 1,002 validation, and 2,005 test images.
\textbf{BloodMNIST}~\cite{medmnistv2} consists of 17,092 microscopy images of peripheral blood cells.
The dataset is officially divided into 11,959 training, 1,712 validation, and 3,421 test samples.
We use the $224\times224$ RGB version for consistency.
\textbf{OrganCMNIST}~\cite{medmnistv2}, derived from LiTS abdominal CT scans, comprises 23,660 coronal-view slices labeled into eleven organ classes.
It includes 13,000 training, 2,392 validation, and 8,268 test images, all resized to $224\times224$.

\subsubsection{Real-world Noisy Datasets}
\textbf{DRTiD}~\cite{hou2022cross} is a real-world diabetic retinopathy dataset comprising 3,100 fundus images collected from community screening, with resolutions ranging from $1,444\times1,444$ to $3,058\times3,000$ pixels.
We split the dataset into 1,600 training, 400 validation, and 1,100 test images.
Initial annotations from community healthcare workers serve as noisy labels, while expert consensus following the International Clinical Diabetic Retinopathy (ICDR) Scale provides the clean ground truth across five severity levels: normal, mild NPDR, moderate NPDR, severe NPDR, and proliferative DR (PDR).
The Kaggle Diabetic Retinopathy (DR) dataset originally contains 88,702 fundus photographs from 44,351 patients (one image per eye) for five-class DR severity classification: normal, mild NPDR, moderate NPDR, severe NPDR, and PDR.
Ju et al.~\cite{ju2022improving} later re-annotated this dataset with assistance from over ten ophthalmologists, releasing the \textbf{Kaggle DR+}, which we adopt in this work to ensure higher annotation quality.
\textbf{CheXpert}~\cite{johnson2019mimic} consists of 224,316 chest radiographs from 65,240 patients, with structured labels automatically extracted from radiology reports.
Among the 14 available observations, we select five clinically important and prevalent conditions: Atelectasis, Cardiomegaly, Consolidation, Edema, and Pleural Effusion.
Following standard practice, uncertain labels are treated as negative, and multi-label samples are removed to retain only single-label cases for training.
The official test set was independently annotated by eight board-certified radiologists, with binarization rules treating present and uncertain likely as positive, and absent and uncertain unlikely as negative.

\subsection{Comparative Study}\label{methodsec}
Our benchmark comprises 10 representative LNL methods together with the \colorbox{cecol}{cross-entropy (CE)} baseline for comparison.
The evaluated methods can be categorized into four groups: noise transition matrix estimation methods (T-Revision~\cite{xia2019anchor} and VolMinNet~\cite{volminnet}), noise-robust regularization methods (SCE~\cite{wang2019symmetric} and CDR~\cite{xia2020robust}), traditional sample selection methods (Co-teaching~\cite{han2018co}, Co-teaching+~\cite{yu2019does}, CoDis~\cite{xia2023combating} and JoCoR~\cite{wei2020combating}), semi-supervised methods (DISC~\cite{li2023disc} and DivideMix~\cite{li2020dividemix}).
\begin{itemize}
  \item \colorbox{matrixcol}{T-Revision}~\cite{xia2019anchor} learns label transition matrices without relying on exact anchor points by introducing a learnable slack variable, enabling more robust classifiers under label noise.
  
\item \colorbox{matrixcol}{VolMinNet}~\cite{volminnet} estimates the label transition matrix without the anchor point assumption by jointly optimizing the cross-entropy loss and the volume of the simplex formed by the matrix columns.

  \item \colorbox{regucol}{SCE}~\cite{wang2019symmetric} combines standard cross-entropy with a reversed cross-entropy loss to tolerate both underfitting on clean data and overfitting on noisy labels.
  
  \item \colorbox{regucol}{CDR}~\cite{xia2020robust} leverages the lottery ticket hypothesis to separate clean-label–fitting and noise-fitting parameters, applying different updates to enhance clean label memorization in early training.

  \item \colorbox{samcol}{Co-teaching}~\cite{han2018co} trains two networks simultaneously and lets each network select small-loss samples to train its peer, reducing the impact of noisy labels.

  \item \colorbox{samcol}{Co-teaching+}~\cite{yu2019does} improves upon Co-teaching by incorporating disagreement-based sample selection, which enhances the robustness of the two networks in extremely noisy scenarios.

  \item \colorbox{samcol}{CoDis}~\cite{xia2023combating} mitigates class imbalance by mining likely-clean samples that exhibit large prediction-probability discrepancies between two jointly trained networks, thereby preserving model divergence.
  
  \item \colorbox{samcol}{JoCoR}~\cite{wei2020combating} jointly optimizes two networks using a shared loss function that balances classification loss and prediction consistency, encouraging agreement on clean samples.

 \item \colorbox{semicol}{DISC}~\cite{li2023disc} dynamically divides training samples into clean, hard, and purified subsets based on two-view consistency, and applies tailored regularization strategies to each subset to enhance robustness against label noise.

  \item \colorbox{semicol}{DivideMix}~\cite{li2020dividemix} fits a Gaussian Mixture Model (GMM) on the training losses to separate clean and noisy samples, then applies semi-supervised learning with MixMatch for robust training.
\end{itemize}

\subsection{Implementation Details}\label{detailsec}
As most existing LNL methods are implemented with ResNet50~\cite{resnet50} as the backbone, we adopt ResNet-50 for fair comparison, while our codebase also supports mainstream backbones, including transformer-based architectures such as ViT~\cite{vit}.
The batch size is set to 128 for PathMNIST, DermaMNIST, BloodMNIST and OrganCMNIST~\cite{medmnistv2}.
For DRTiD~\cite{hou2022cross} and Kaggle DR+~\cite{ju2022improving}, all fundus images are resized to $512\times512$ before training, and the batch size is reduced to 16.
For CheXpert~\cite{johnson2019mimic}, all fundus images are resized to $224\times224$ before training, and the batch size is 128.
For each method, we adopt the optimizer and learning rate settings recommended in their original methods to preserve their optimal training configurations. 
All experiments are conducted on a single machine equipped with four NVIDIA A6000 GPUs. 
To ensure reproducibility, each experiment is repeated with three different random seeds, and the averaged results are reported.

\subsection{Experimental Result Overview}\label{overallsec}
To obtain a stable and fair comparison, we report the average classification accuracy over the last 5 epochs for each method under three noise patterns: symmetric noise, instance-dependent noise, and real-world noise. 
Based on these results, we rank all methods according to their average performance, and further compute the mean accuracy across the three noise patterns to provide an overall measure of robustness~(see Fig.~\ref{fig:teaser}).
Overall, Co-teaching and DISC  achieved the highest average classification accuracy under all noise patterns. 
Co-teaching+ and DivideMix achieved the highest average classification accuracy under symmetric noise.
Co-teaching and CoDis achieved the highest average classification accuracy under instance-dependent noise.
The noise-robust regularization methods~(SCE and CDR) achieved the highest average classification accuracy under real-world noise.

\subsection{Results on Balanced Benchmark Dataset under Synthetic Noise}\label{balancesec}
In this section, we evaluate LNL methods on the relatively balanced PathMNIST, considering both symmetric noise and instance-dependent noise.

\subsubsection{Quantitative Results}
Table~\ref{pathmnist_results} summarizes the quantitative results of 10 representative LNL methods on the PathMNIST under symmetric and instance-dependent noise settings, with 3 different noise rates: 20\%, 50\%, and 90\%. 

All methods exhibit good  performance under Sym-20\%.
Under Sym-50\%, both transition matrix based~(T-revision and VolMinNet) and regularization methods~(SCE and CDR) exhibit performance degradation.
In particular, we find a substantial discrepancy between the V and L accuracies for these methods, indicating a tendency to overfit noisy labels in the later stages of training.
Most sample selection and semi-supervised methods continue to perform well under Sym-50\%, with the exception of JoCoR.
At Sym-90\%, all methods experience significant performance degradation.

In the more realistic and challenging instance-dependent noise setting, performance declines significantly across all methods as the noise rate increases. 
At Idn-20\%, most methods still retain reasonable performance.
As the noise rate increases to 50\%, noise-robust regularization methods  and noise transition matrix estimation methods begin to show performance degradation, while Co-teaching, CoDis and semi-supervised methods achieve higher accuracy.
At Idn-90\%, all methods exhibit a complete collapse in performance. 

In summary, methods relying on regularization or noise transition matrix estimation exhibit limited effectiveness. 
In contrast, DISC, DivideMix, CoDis and Co-teaching demonstrate comparatively superior robustness.
In this section, we presented a quantitative comparison of different methods.
Next, we delves into a deeper analysis to understand the underlying reasons behind their performance differences.

\begin{figure}
    \centering
    \begin{subfigure}[b]{0.48\columnwidth}
        \centering
        \includegraphics[width=\columnwidth]{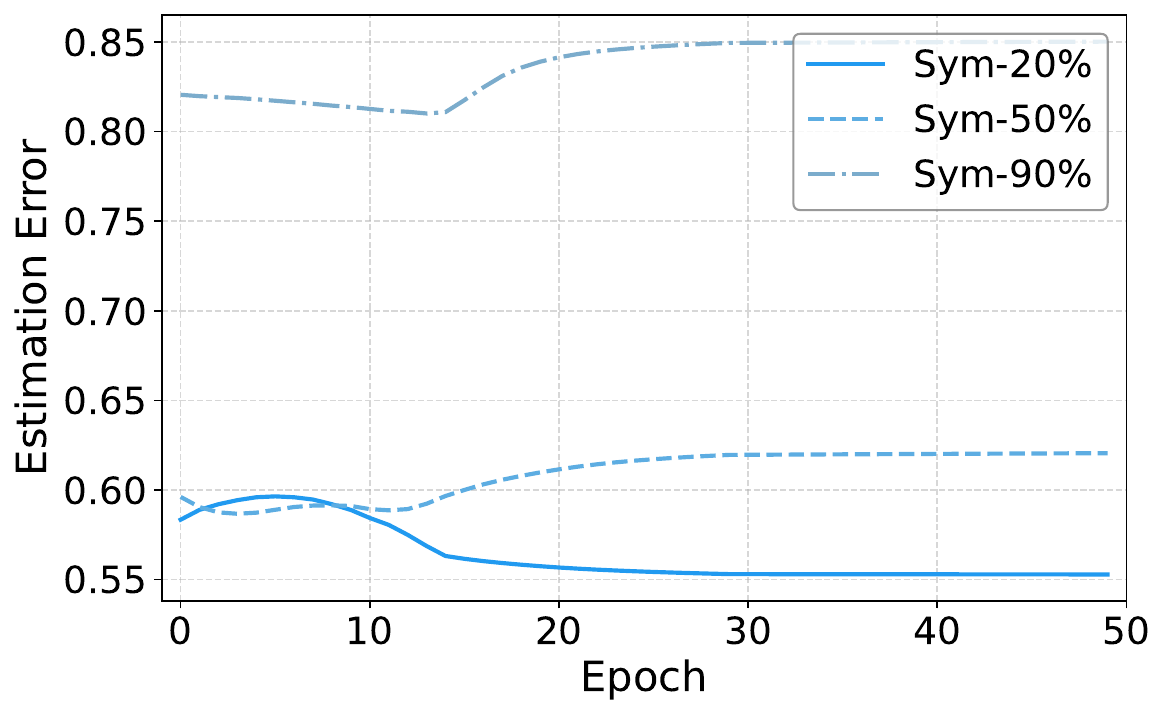}
        \caption{Estimation Error}
        \label{fig:es}
    \end{subfigure}
    \hfill
    \begin{subfigure}[b]{0.48\columnwidth}
        \centering
        \includegraphics[width=\columnwidth]{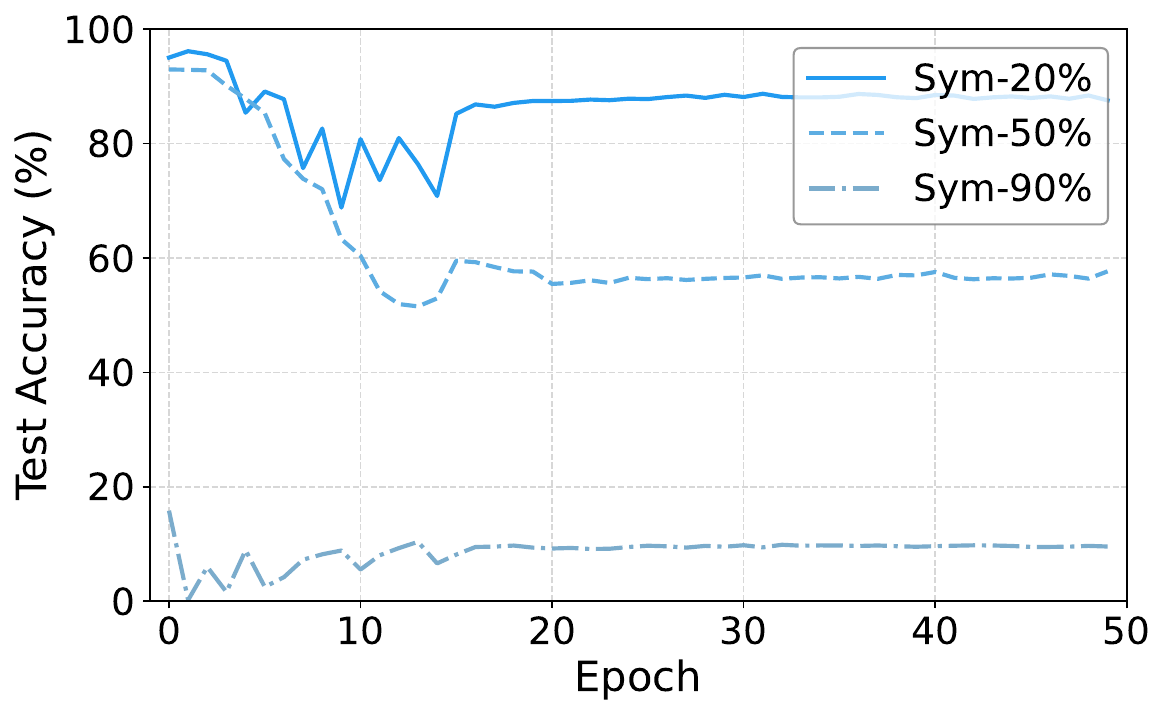}
        \caption{Test Accuracy}
        \label{fig:acc}
    \end{subfigure}

    \caption{Comparison of estimation error and test accuracy under different symmetric noise rates for VolMinNet.}
    \label{matrix_er}
\end{figure}
 
\subsubsection{Analysis on Noise Transition Matrix Estimation} 
Transition matrix estimation methods exhibit a substantial performance gap between symmetric and instance-dependent noise.
Since these methods are specifically designed for instance-independent noise.
For symmetric noise (see Fig.~\ref{matrix_er}), we observe that for VolMinNet, smaller estimation error of the transition matrix is usually associated with higher test accuracy. 
This indicates a strong correlation between estimation quality and classification performance. 
However, when the noise rate exceeds 50\%, the estimation error grows significantly, resulting in a noticeable drop in classification accuracy.
This demonstrates that these methods fail to resolve the label noise issue, even on balanced dataset.


\begin{figure}
  \centering

  \begin{subfigure}{0.48\columnwidth}
    \centering
    \includegraphics[width=\linewidth]{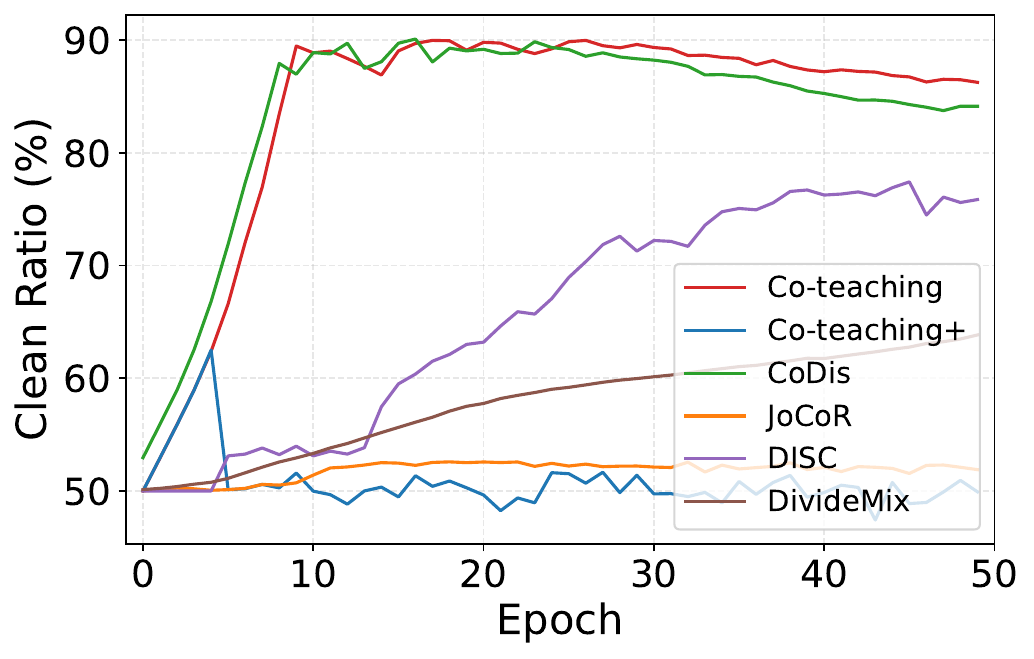}
    \caption{}
    \label{fig:samplepathmnista}
  \end{subfigure}
  \hfill
  \begin{subfigure}{0.48\columnwidth}
    \centering
    \includegraphics[width=\linewidth]{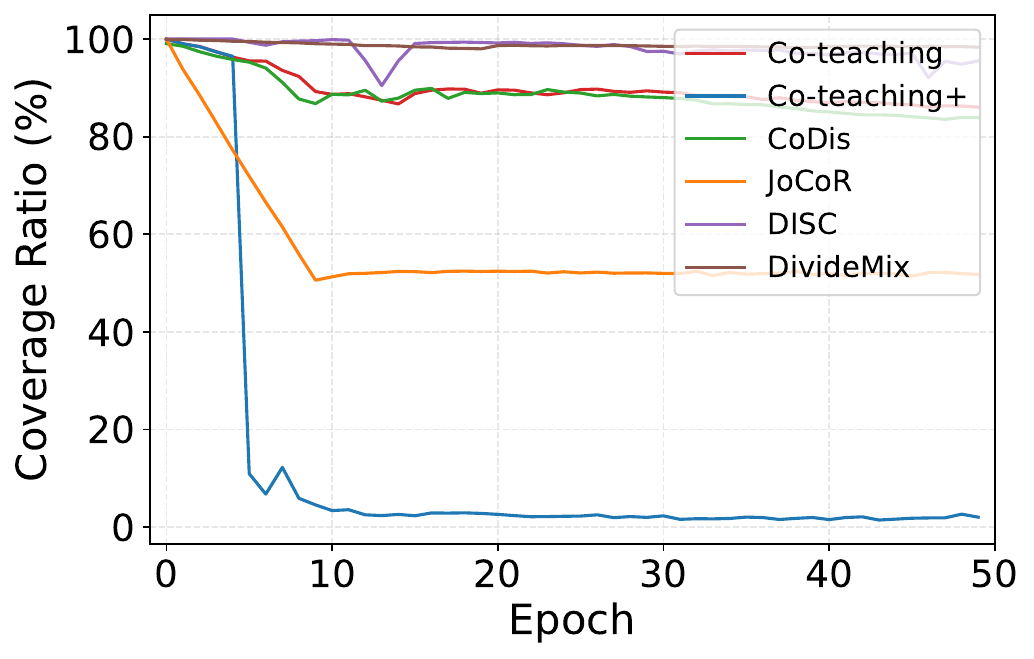}
    \caption{}
    \label{fig:samplepathmnistb}
  \end{subfigure}

  \vspace{0.4em}
  \begin{subfigure}{0.48\columnwidth}
    \centering
    \includegraphics[width=\linewidth]{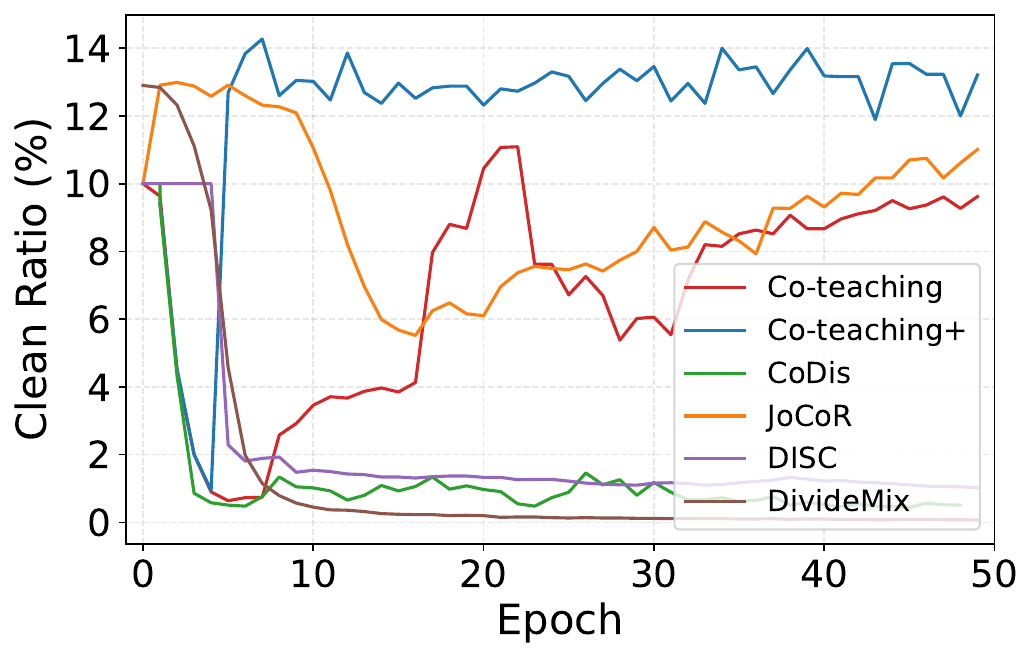}
    \caption{}
    \label{fig:samplepathmnistc}
  \end{subfigure}
  \hfill
  \begin{subfigure}{0.48\columnwidth}
    \centering
    \includegraphics[width=\linewidth]{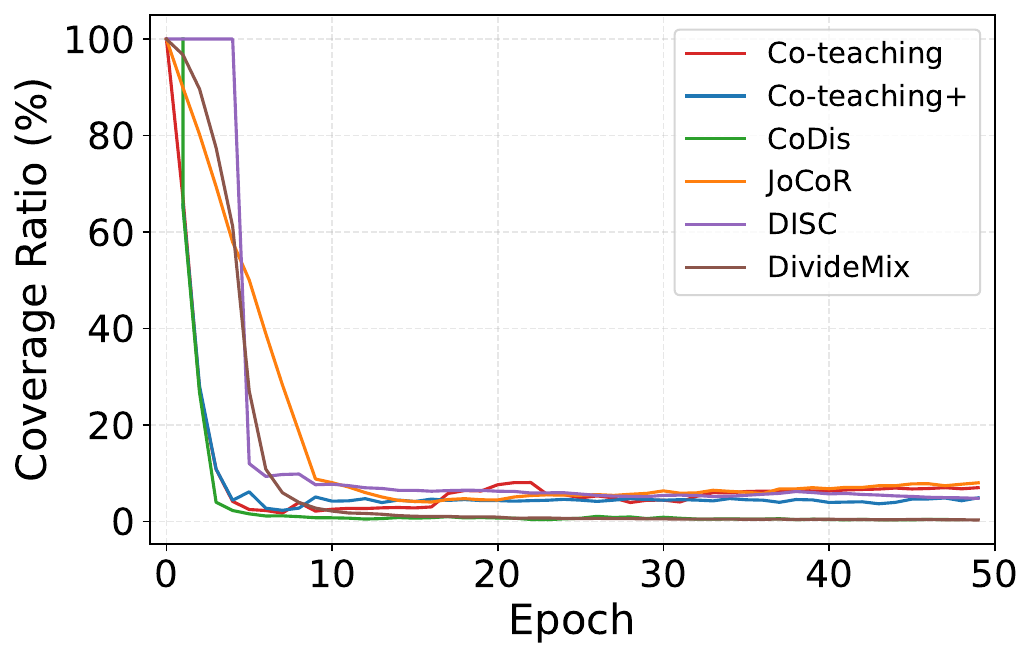}
    \caption{}
    \label{fig:samplepathmnistd}
  \end{subfigure}

  \caption{Sample selection performance under instance-dependent noise on PathMNIST. 
  The clean ratio represents the proportion of correctly labeled samples among those selected, 
  while the coverage ratio represents the proportion of selected clean samples among all clean samples. 
  (a) Clean ratio at 50\% noise. 
  (b) Coverage ratio at 50\% noise. 
  (c) Clean ratio at 90\% noise. 
  (d) Coverage ratio at 90\% noise.}
  \label{fig:samplepathmnist}
\end{figure}

\begin{figure}
    \centering

    \begin{subfigure}[b]{0.48\columnwidth}
        \centering
        \includegraphics[width=\columnwidth]{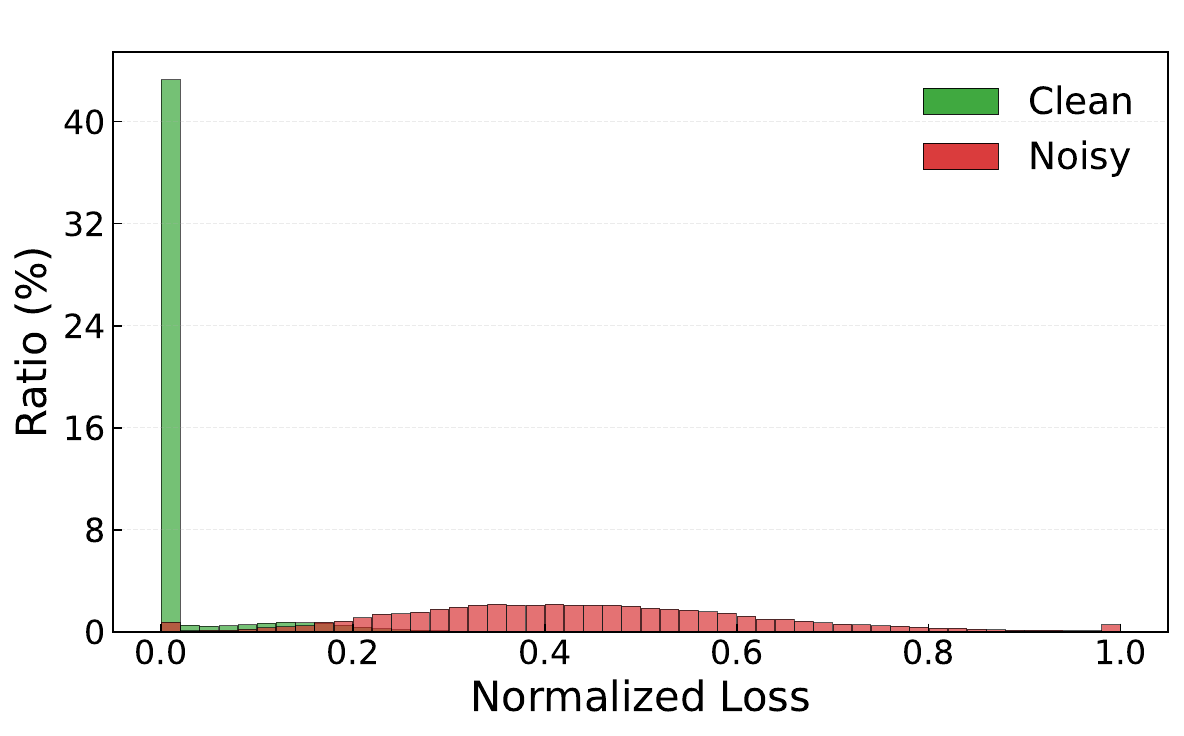}
        \caption{Idn-50\%}
        \label{fig:copa50}
    \end{subfigure}
    \hfill
    \begin{subfigure}[b]{0.48\columnwidth}
        \centering
        \includegraphics[width=\columnwidth]{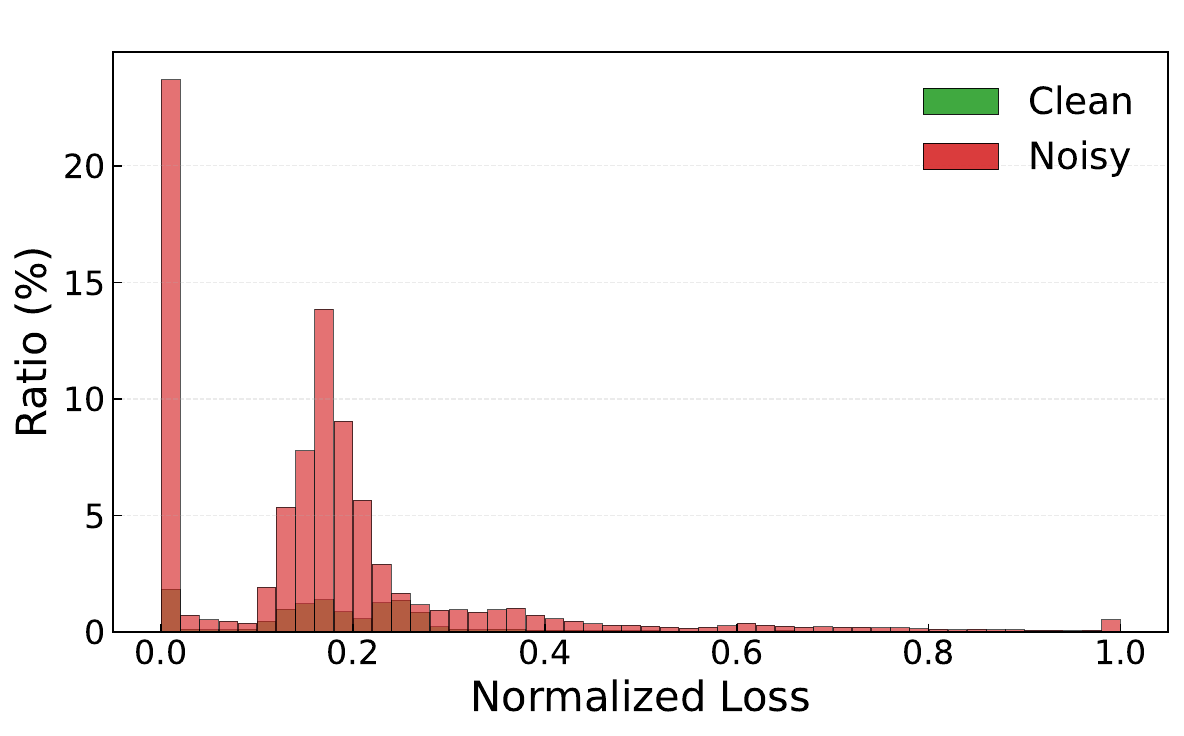}
        \caption{Idn-90\%}
        \label{fig:copa90}
    \end{subfigure}

    \caption{The loss distributions of Co-teaching under Idn-50\% and Idn-90\% on PathMNIST.}
    \label{copa}
\end{figure}

\begin{table*}[t]
\caption{Comparison of different methods on three imbalance datasets under instance-dependent noise. \textbf{Bold} indicates the best result}
\label{derma_multi}
\scriptsize
\centering
\begin{tblr}{
  cell{1}{3} = {c=3}{},
  cell{1}{6} = {c=3}{},
  cell{1}{9} = {c=3}{},
  cell{3}{1}  = {r=3}{},
  cell{6}{1}  = {r=3}{},
  cell{9}{1}  = {r=3}{},
  cell{12}{1} = {r=3}{},
  cell{15}{1} = {r=3}{},
  cell{18}{1} = {r=3}{},
  cell{21}{1} = {r=3}{},
  cell{24}{1} = {r=3}{},
  cell{27}{1} = {r=3}{},
  cell{30}{1} = {r=3}{},
  cell{33}{1} = {r=3}{},
  colspec={c c *{9}{c}},
  rows  = {abovesep=0.3ex, belowsep=0.3ex},
  colsep = 3pt,
  row{1} = {font=\bfseries, halign=c},
  hline{1,2,3,Z} = {-}{},
  row{3-5}   = {bg=cecol},
  row{6-11}  = {bg=matrixcol},
  row{12-17} = {bg=regucol},
  row{18-29} = {bg=samcol},
  row{30-35} = {bg=semicol},
}
Dataset &        & \SetCell[c=3]{c}DermaMNIST & & & \SetCell[c=3]{c}BloodMNIST & & & \SetCell[c=3]{c}OrganCMNIST & & & \\
Noise type &     & Idn-20\% & Idn-50\% & Idn-90\% & Idn-20\% & Idn-50\% & Idn-90\% & Idn-20\% & Idn-50\% & Idn-90\% \\
CE
  & B & 80.60 $\pm$ 1.15 & 69.63 $\pm$ 3.18 & 22.15 $\pm$ 1.67 & 94.68 $\pm$ 1.35 & 75.15 $\pm$ 1.81 & 17.36 $\pm$ 1.75 & 90.21 $\pm$ 0.98 & 68.17 $\pm$ 1.49 & 27.56 $\pm$ 2.79 \\
  & V & 80.55 $\pm$ 1.57 & 66.53 $\pm$ 4.36 & \textbf{17.11 $\pm$ 1.41} & 94.53 $\pm$ 0.48 & 75.15 $\pm$ 1.13 & 2.81 $\pm$ 2.91 & 89.28 $\pm$ 0.97 & 68.13 $\pm$ 1.46 & 15.01 $\pm$ 2.78 \\
  & L & 79.97 $\pm$ 0.33 & 55.43 $\pm$ 0.47 & 16.68 $\pm$ 1.20 & 93.26 $\pm$ 0.86 & 63.79 $\pm$ 1.78 & 7.71 $\pm$ 3.28 & 88.06 $\pm$ 0.78 & 65.80 $\pm$ 1.75 & 14.94 $\pm$ 3.87 \\
  \midrule
T-Revision
  & B & 73.16 $\pm$ 1.23 & 35.66 $\pm$ 1.61 & 18.65 $\pm$ 1.19 & 96.61 $\pm$ 0.95 & 79.77 $\pm$ 1.05 & 17.36 $\pm$ 1.75 & 89.42 $\pm$ 0.71 & 66.97 $\pm$ 1.35 & 17.12 $\pm$ 1.35 \\
  & V & 71.77 $\pm$ 1.08 & 35.14 $\pm$ 3.71 & 16.08 $\pm$ 1.39 & 93.62 $\pm$ 0.82 & 62.41 $\pm$ 1.63 & 11.64 $\pm$ 2.71 & 89.34 $\pm$ 0.84 & 65.74 $\pm$ 1.85 & 14.82 $\pm$ 1.73 \\
  & L & 72.47 $\pm$ 0.56 & 17.57 $\pm$ 0.79 & 14.17 $\pm$ 1.94 & 93.10 $\pm$ 0.47 & 57.14 $\pm$ 0.65 & 18.15 $\pm$ 2.96 & 89.14 $\pm$ 0.49 & 60.56 $\pm$ 0.64 & 16.33 $\pm$ 2.18 \\
   \midrule
VolMinNet
  & B & 72.74 $\pm$ 1.52 & 39.63 $\pm$ 3.24 & 19.48 $\pm$ 2.13 & 96.17 $\pm$ 1.46 & 72.70 $\pm$ 1.51 & 19.64 $\pm$ 2.19 & 88.33 $\pm$ 0.93 & 68.79 $\pm$ 1.09 & 24.22 $\pm$ 2.82 \\
  & V & 71.24 $\pm$ 1.48 & 34.99 $\pm$ 2.74 & 15.71 $\pm$ 1.65 & 96.12 $\pm$ 0.95 & 72.71 $\pm$ 1.47 & 8.71 $\pm$ 3.24 & 88.33 $\pm$ 0.62 & 60.03 $\pm$ 2.08 & 13.13 $\pm$ 1.95 \\
  & L & 71.23 $\pm$ 0.79 & 38.14 $\pm$ 1.17 & 15.66 $\pm$ 1.38 & 91.33 $\pm$ 1.13 & 61.53 $\pm$ 1.36 & 8.98 $\pm$ 2.19 & 87.36 $\pm$ 0.72 & 63.46 $\pm$ 1.68 & 11.91 $\pm$ 2.97 \\
  \midrule
SCE
  & B & 82.18 $\pm$ 0.58 & 71.50 $\pm$ 0.36 & 23.08 $\pm$ 0.71 & 97.12 $\pm$ 1.33 & 86.84 $\pm$ 0.67 & 20.41 $\pm$ 1.22 & 93.17 $\pm$ 0.16 & 81.89 $\pm$ 1.00 & 23.27 $\pm$ 0.69 \\
  & V & \textbf{82.54 $\pm$ 0.79} & 67.81 $\pm$ 0.95 & 13.80 $\pm$ 0.53 & 96.02 $\pm$ 1.36 & 76.21 $\pm$ 1.63 & 7.01 $\pm$ 1.62 & 92.21 $\pm$ 1.20 & 80.14 $\pm$ 3.38 & 11.70 $\pm$ 0.96 \\
  & L & 77.35 $\pm$ 0.81 & 57.20 $\pm$ 0.73 & 18.67 $\pm$ 0.86 & 89.62 $\pm$ 0.75 & 50.95 $\pm$ 0.92 & \textbf{19.02 $\pm$ 0.16} & 86.06 $\pm$ 0.68 & 55.91 $\pm$ 0.35 & 21.65 $\pm$ 0.25 \\
  \midrule
CDR
  & B & \textbf{82.39 $\pm$ 0.86} & 71.52 $\pm$ 0.57 & \textbf{32.62 $\pm$ 0.85} & 96.34 $\pm$ 0.26 & 76.38 $\pm$ 0.83 & 15.16 $\pm$ 0.89 & 91.95 $\pm$ 0.16 & 75.87 $\pm$ 1.19 & 24.56 $\pm$ 2.56 \\
  & V & 82.14 $\pm$ 0.71 & 67.13 $\pm$ 0.34 & 11.87 $\pm$ 0.49 & 95.14 $\pm$ 0.54 & 77.27 $\pm$ 0.43 & 7.97 $\pm$ 0.92 & 90.01 $\pm$ 1.63 & 74.85 $\pm$ 1.01 & 16.76 $\pm$ 1.39 \\
  & L & \textbf{81.95 $\pm$ 0.78} & 57.77 $\pm$ 1.14 & 15.80 $\pm$ 0.76 & 95.97 $\pm$ 1.03 & 65.43 $\pm$ 0.79 & 9.32 $\pm$ 0.56 & 91.68 $\pm$ 0.18 & 67.97 $\pm$ 0.40 & 15.63 $\pm$ 0.19 \\
  \midrule
Co-teaching
  & B & 74.36 $\pm$ 1.41 & 70.07 $\pm$ 0.79 & 20.55 $\pm$ 0.85 & 96.34 $\pm$ 0.19 & 89.27 $\pm$ 2.55 & 17.96 $\pm$ 7.90 & 91.25 $\pm$ 0.07 & 82.64 $\pm$ 1.14 & 11.11 $\pm$ 2.44 \\
  & V & 73.99 $\pm$ 0.89 & 66.23 $\pm$ 1.29 & 11.65 $\pm$ 1.23 & 95.99 $\pm$ 0.44 & 88.68 $\pm$ 3.09 & \textbf{11.65 $\pm$ 1.23} & 90.83 $\pm$ 0.36 & 81.92 $\pm$ 1.09 & \textbf{21.20 $\pm$ 0.32} \\
  & L & 73.08 $\pm$ 0.48 & 65.65 $\pm$ 0.77 & 11.54 $\pm$ 0.47 & 95.79 $\pm$ 0.35 & 86.15 $\pm$ 2.12 & 7.47 $\pm$ 0.94 & 90.87 $\pm$ 0.07 & \textbf{82.09 $\pm$ 1.12} & 14.84 $\pm$ 2.55 \\
  \midrule
Co-teaching+
  & B & 72.19 $\pm$ 0.28 & 67.93 $\pm$ 4.18 & 15.28 $\pm$ 3.16 & 94.74 $\pm$ 1.87 & 77.17 $\pm$ 2.00 & \textbf{21.40 $\pm$ 5.76} & 88.92 $\pm$ 0.71 & 73.95 $\pm$ 1.46 & \textbf{30.78 $\pm$ 6.81} \\
  & V & 72.17 $\pm$ 0.36 & 62.74 $\pm$ 5.47 & 11.59 $\pm$ 2.55 & 94.69 $\pm$ 1.87 & 75.52 $\pm$ 0.89 & 4.74 $\pm$ 0.18 & 88.33 $\pm$ 0.90 & 73.91 $\pm$ 1.46 & 11.63 $\pm$ 2.90 \\
  & L & 66.88 $\pm$ 0.57 & 61.17 $\pm$ 7.11 & 10.97 $\pm$ 1.46 & 91.82 $\pm$ 3.32 & 54.64 $\pm$ 5.82 & 18.18 $\pm$ 5.70 & 88.29 $\pm$ 0.99 & 60.84 $\pm$ 0.97 & \textbf{23.20 $\pm$ 8.56} \\
  \midrule
CoDis
  & B & 74.71 $\pm$ 0.95 & 69.78 $\pm$ 1.34 & 18.68 $\pm$ 2.12 & 95.81 $\pm$ 0.10 & 91.79 $\pm$ 0.99 & 14.20 $\pm$ 3.44 & 90.94 $\pm$ 0.38 & 82.45 $\pm$ 0.13 & 22.34 $\pm$ 5.16 \\
  & V & 74.64 $\pm$ 0.56 & 67.01 $\pm$ 2.68 & 18.59 $\pm$ 1.17 & 95.61 $\pm$ 0.25 & 88.04 $\pm$ 1.97 & 5.62 $\pm$ 2.69 & 90.06 $\pm$ 1.09 & 78.64 $\pm$ 2.15 & 9.57 $\pm$ 0.60 \\
  & L & 72.81 $\pm$ 0.47 & 61.31 $\pm$ 2.35 & 13.15 $\pm$ 1.66 & 95.34 $\pm$ 0.09 & 87.30 $\pm$ 1.48 & 10.70 $\pm$ 2.87 & 90.60 $\pm$ 0.36 & 80.47 $\pm$ 0.53 & 7.65 $\pm$ 2.37 \\
  \midrule
JoCoR
  & B & 74.14 $\pm$ 1.19 & 66.76 $\pm$ 1.37 & 24.59 $\pm$ 3.64 & 92.86 $\pm$ 0.51 & 71.47 $\pm$ 3.94 & 20.82 $\pm$ 2.56 & 87.96 $\pm$ 0.57 & 75.04 $\pm$ 0.71 & 26.47 $\pm$ 2.84 \\
  & V & 73.04 $\pm$ 1.24 & 66.41 $\pm$ 2.18 & 16.83 $\pm$ 2.81 & 92.68 $\pm$ 0.62 & 70.52 $\pm$ 4.41 & 9.06 $\pm$ 1.58 & 87.57 $\pm$ 0.99 & 74.43 $\pm$ 0.60 & 17.12 $\pm$ 1.58 \\
  & L & 71.78 $\pm$ 0.79 & 53.69 $\pm$ 0.92 & \textbf{20.82 $\pm$ 1.12} & 92.14 $\pm$ 0.23 & 63.94 $\pm$ 0.62 & 10.69 $\pm$ 0.64 & 87.56 $\pm$ 0.27 & 63.75 $\pm$ 1.19 & 18.97 $\pm$ 0.39 \\
  \midrule
DISC
  & B & 80.35 $\pm$ 0.78 & \textbf{71.97 $\pm$ 0.58} & 23.69 $\pm$ 2.11 & \textbf{98.74 $\pm$ 0.19} & \textbf{93.24 $\pm$ 1.34} & 16.83 $\pm$ 6.95 & 89.90 $\pm$ 2.75 & 82.67 $\pm$ 0.86 & 25.00 $\pm$ 3.66 \\
  & V & 80.33 $\pm$ 1.13 & \textbf{68.73 $\pm$ 2.14} & 15.16 $\pm$ 3.78 & \textbf{98.41 $\pm$ 0.47} & 89.19 $\pm$ 0.52 & 1.91 $\pm$ 0.25 & 88.68 $\pm$ 2.97 & 80.66 $\pm$ 2.14 & 7.81 $\pm$ 1.06 \\
  & L & 79.49 $\pm$ 1.57 & \textbf{68.19 $\pm$ 1.15} & 10.97 $\pm$ 1.19 & \textbf{98.59 $\pm$ 0.26} & \textbf{88.83 $\pm$ 0.82} & 1.76 $\pm$ 0.41 & 88.99 $\pm$ 3.14 & 80.59 $\pm$ 0.14 & 8.14 $\pm$ 1.33 \\
  \midrule
DivideMix
  & B & 74.96 $\pm$ 0.83 & 66.93 $\pm$ 0.26 & 29.75 $\pm$ 1.85 & 98.32 $\pm$ 0.23 & 84.99 $\pm$ 8.46 & 15.94 $\pm$ 4.45 & \textbf{93.20 $\pm$ 0.05} & \textbf{83.47 $\pm$ 0.50} & 22.05 $\pm$ 0.12 \\
  & V & 72.37 $\pm$ 1.14 & 66.83 $\pm$ 0.41 & 14.76 $\pm$ 2.13 & 98.06 $\pm$ 0.37 & \textbf{90.46 $\pm$ 3.64} & 3.82 $\pm$ 0.48 & \textbf{92.58 $\pm$ 0.09} & \textbf{82.25 $\pm$ 1.36} & 15.26 $\pm$ 1.48 \\
  & L & 66.72 $\pm$ 0.39 & 66.86 $\pm$ 0.77 & 13.19 $\pm$ 3.38 & 97.58 $\pm$ 0.62 & 81.16 $\pm$ 3.74 & 5.42 $\pm$ 4.77 & \textbf{92.44 $\pm$ 0.24} & 76.78 $\pm$ 0.90 & 6.73 $\pm$ 0.71 \\
\end{tblr}
\end{table*}

\subsubsection{Analysis on Sample Selection}
Next, we focus on sample selection and semi-supervised methods, which have been effective under instance dependent noise.
The sample selection methods in LNMBench are all derived from the Co-teaching, adopting the small-loss strategy in which the proportion of selected samples is determined by the known noise rate.
Next, we take the Co-teaching method as an example to analyze how the small-loss strategy performs under varying noise rates.
As shown in Fig.~\ref{copa},  most clean samples exhibit lower losses than noisy ones, making it feasible to distinguish them via the small-loss criterion under Idn-50\%.
However, when the noise rate increases to 90\%, the model rapidly overfit noisy labels, causing noisy samples to also present small losses.
Consequently, as shown in Fig~\ref{fig:samplepathmnist}, under Idn-50\%, Co-teaching is able to select training samples with a high clean ratio while retaining nearly all clean samples, which leads to relatively high classification accuracy. 
In contrast, under Idn-90\%, the method fails to identify clean samples, resulting in significant performance degradation.

Under Idn-50\% on the PathMNIST, we observed that a considerable proportion of noisy samples were still included in the clean set of DISC and DivideMix. 
They also maintain a reasonably good classification accuracy, which indicates that the semi-supervised strategy can enhance the model’s robustness to label noise.
For Disc, this phenomenon mainly arises because some noisy samples, once incorrectly fitted by the model, gradually achieve higher prediction scores on their observed labels. 
As the confidence threshold is dynamically updated during training, these scores eventually exceed the selection criterion, leading to their erroneous inclusion in the clean set.
DivideMix also adopts the small-loss strategy but does not use the noise rate as a threshold, which leads to fitting part of the noisy labels during the selection process. 
As a result, the clean set inevitably contains some noisy samples.

\textbf{Findings:} Noise-robust regularization methods and transition matrix estimation methods are only effective under low noise ratio.
For PathMNIST with known noise rates, Co-teaching and CoDis, which based on small loss selection shows superior capability in identifying clean samples.
Semi-supervised methods do not rely on the known noise rate for sample selection, and therefore their selection ability is less effective, it effectively mitigates the impact of noisy labels through its semi-supervised learning strategy.
However, in real-world scenarios the noise rate is typically unknown, which makes such semi-supervised methods more practical.

\subsection{Results on Imbalanced Datasets under Synthetic Noise}\label{imbalancesec}
This section examines how class imbalance affects the robustness of LNL methods (see Table~\ref{derma_multi}). 
We evaluate on three class imbalanced medical imaging datasets with instance-dependent noise (DermaMNIST, BloodMNIST, and OrganCMNIST.
We omit symmetric noise in this section, as it is less realistic in practice.

\subsubsection{Quantitative Results}
For BloodMNIST, all methods attain high accuracy under Idn-20\%. 
Under Idn-50\%, only Co-teaching, CoDis, and semi-supervised methods achieve high accuracy. 
Under Idn-90\%, all methods deteriorate to very low accuracy. 
These methods exhibit the same trend on OrganCMNIST as on BloodMNIST.
For DermaMNIST, CDR achieves the highest accuracy under the Idn-20\% setting, but suffers a substantial performance drop at Idn-50\%.
Under Idn-50\%, DISC attains the best results, followed closely by DivideMix and Co-teaching.
Under Idn-90\%, all methods fail to maintain satisfactory performance.
Notably, under Idn-20\%, both sample selection and semi-supervised methods perform worse than the standard CE baseline, whereas at Idn-50\% they surpass CE.

\begin{figure}
    \centering

    \begin{subfigure}[b]{0.48\columnwidth}
        \centering
        \includegraphics[width=\columnwidth]{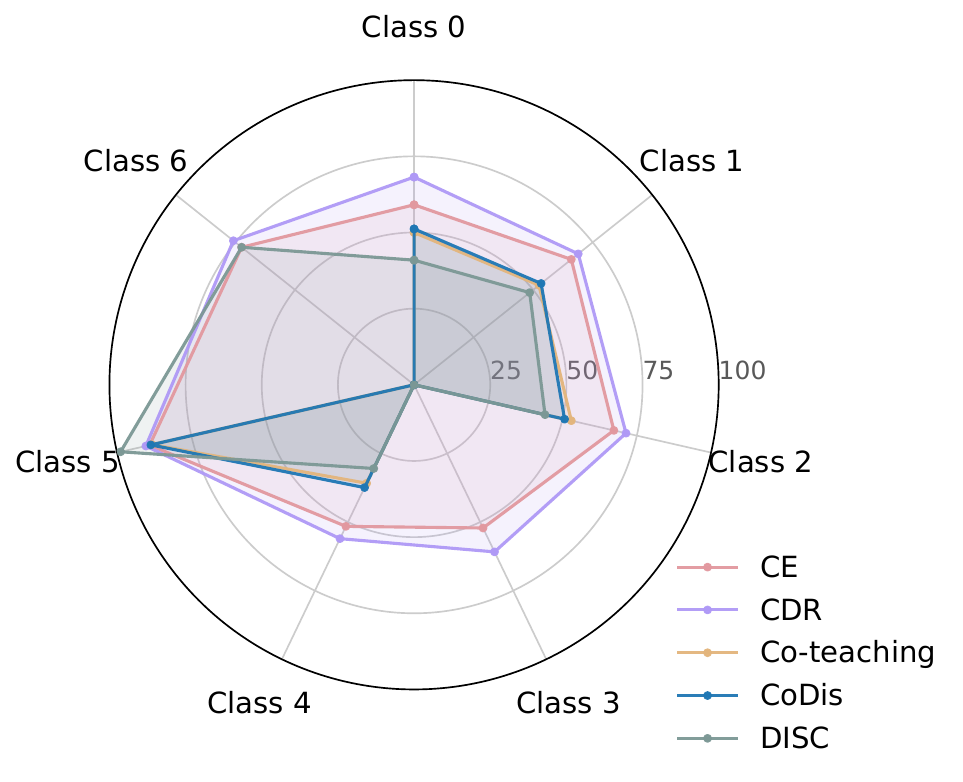}
        \caption{Idn20\%}
        \label{fig:derma20}
    \end{subfigure}
    \hfill
    \begin{subfigure}[b]{0.48\columnwidth}
        \centering
        \includegraphics[width=\columnwidth]{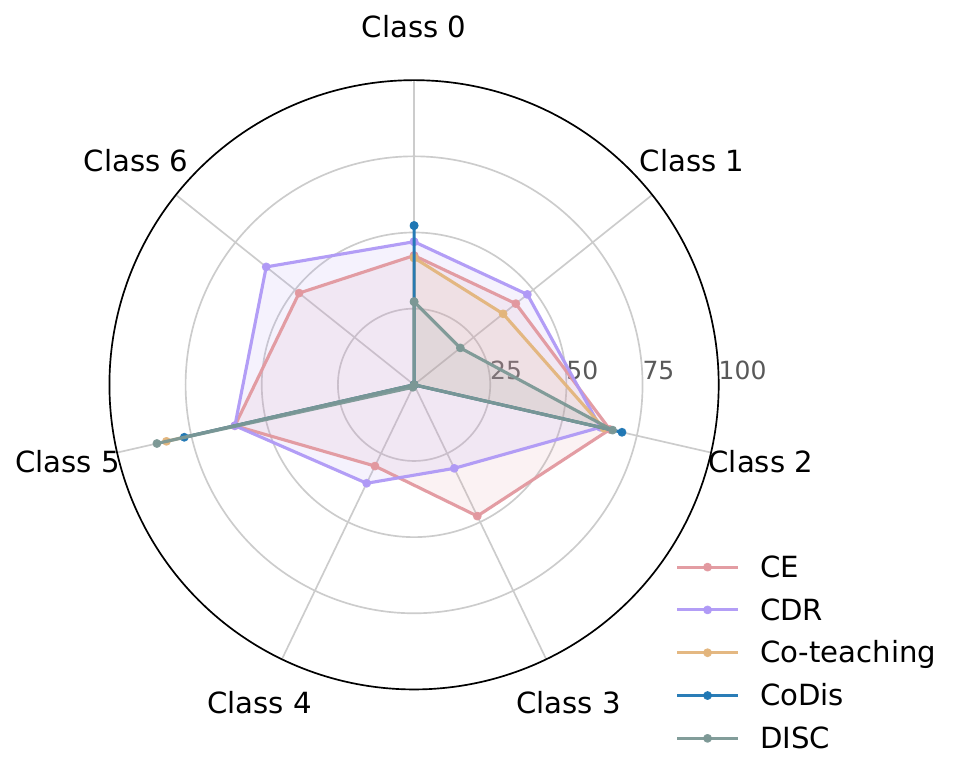}
        \caption{Idn50\%}
        \label{fig:derma50}
    \end{subfigure}

    \caption{Per-class accuracy of representative methods under Idn-20\% and Idn-50\% on DermaMNIST.}
    \label{derma}
\end{figure}

To further investigate the underlying causes of this performance reversal on DermaMNIST, we present per-class accuracy results for four representative methods and CE~(see Fig.~\ref{fig:derma20}).
At Idn-20\% setting, CDR consistently outperforms CE across all classes.
In contrast, DISC, CoDis and Co-teaching underperform CE in most classes except for class 5, which is the majority class in DermaMNIST.
At Idn-50\%, CDR exhibits performance degradation across all classes, further confirming that regularization based methods fail under high noise levels (see Fig.~\ref{fig:derma50}).
Meanwhile, DISC, CoDis and Co-teaching still underperform CE in most classes except for class 5.
These results suggest that sample selection based methods are ineffective on the DermaMNIST.




\begin{figure}
  \centering

  \begin{subfigure}{0.48\columnwidth}
    \centering
    \includegraphics[width=\linewidth]{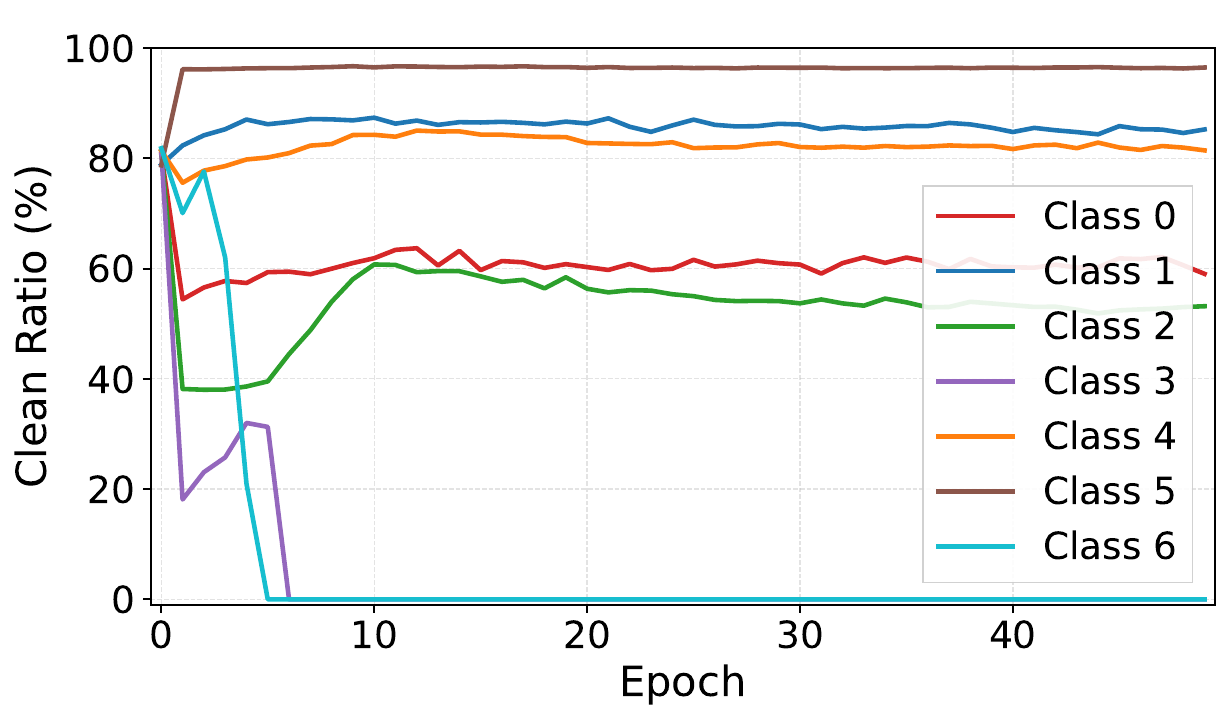}
    \caption{}
    \label{fig:precision_derco}
  \end{subfigure}
  \hfill
  \begin{subfigure}{0.48\columnwidth}
    \centering
    \includegraphics[width=\linewidth]{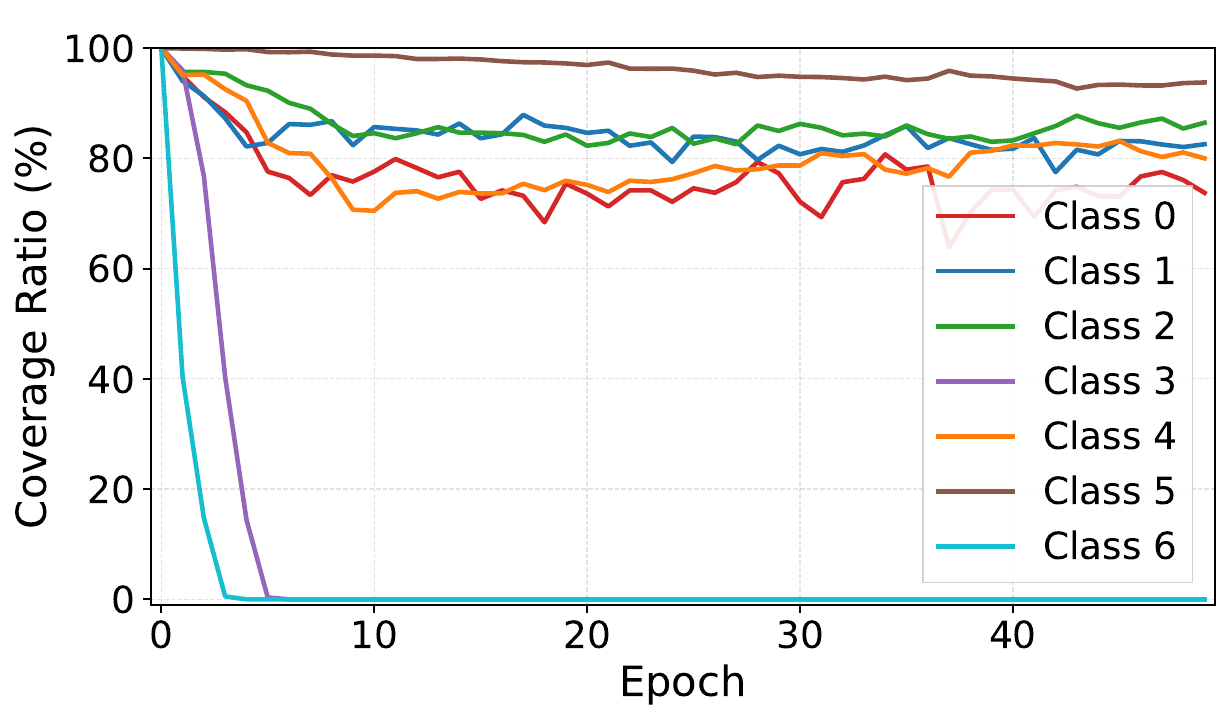}
    \caption{}
    \label{fig:recall_derco}
  \end{subfigure}

  \vspace{0.5em}
  \begin{subfigure}{0.48\columnwidth}
    \centering
    \includegraphics[width=\linewidth]{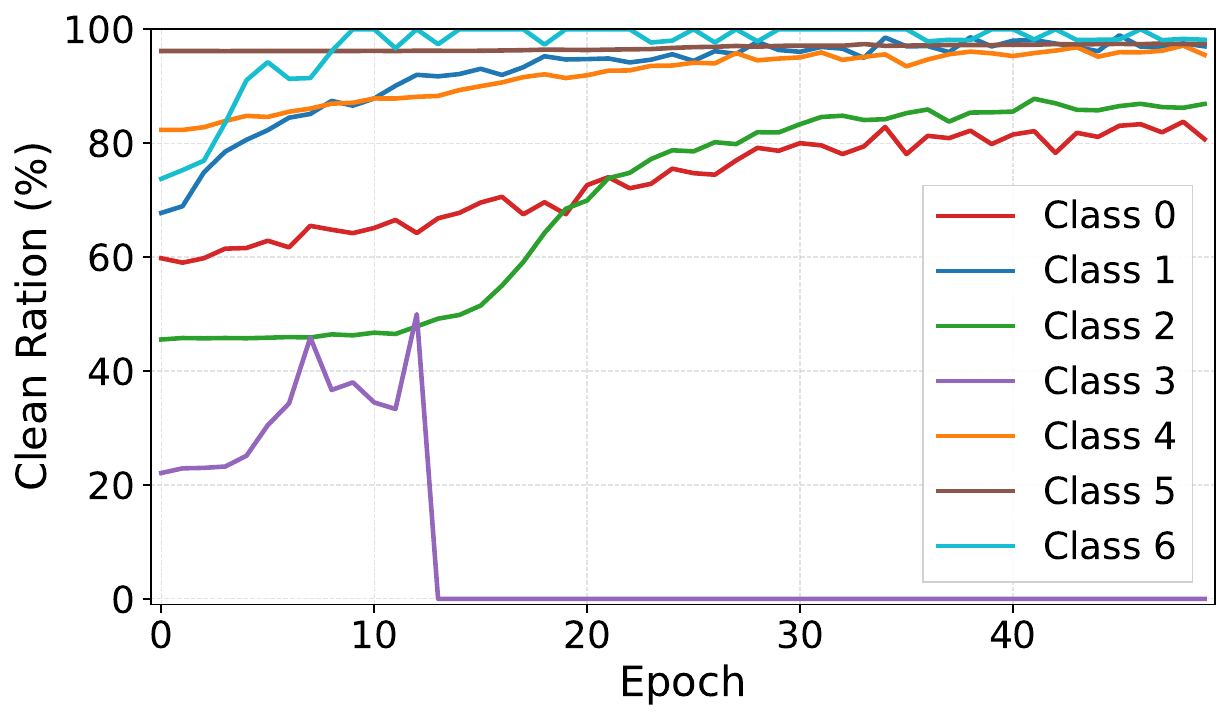}
    \caption{}
    \label{fig:precision_derdisc}
  \end{subfigure}
  \hfill
  \begin{subfigure}{0.48\columnwidth}
    \centering
    \includegraphics[width=\linewidth]{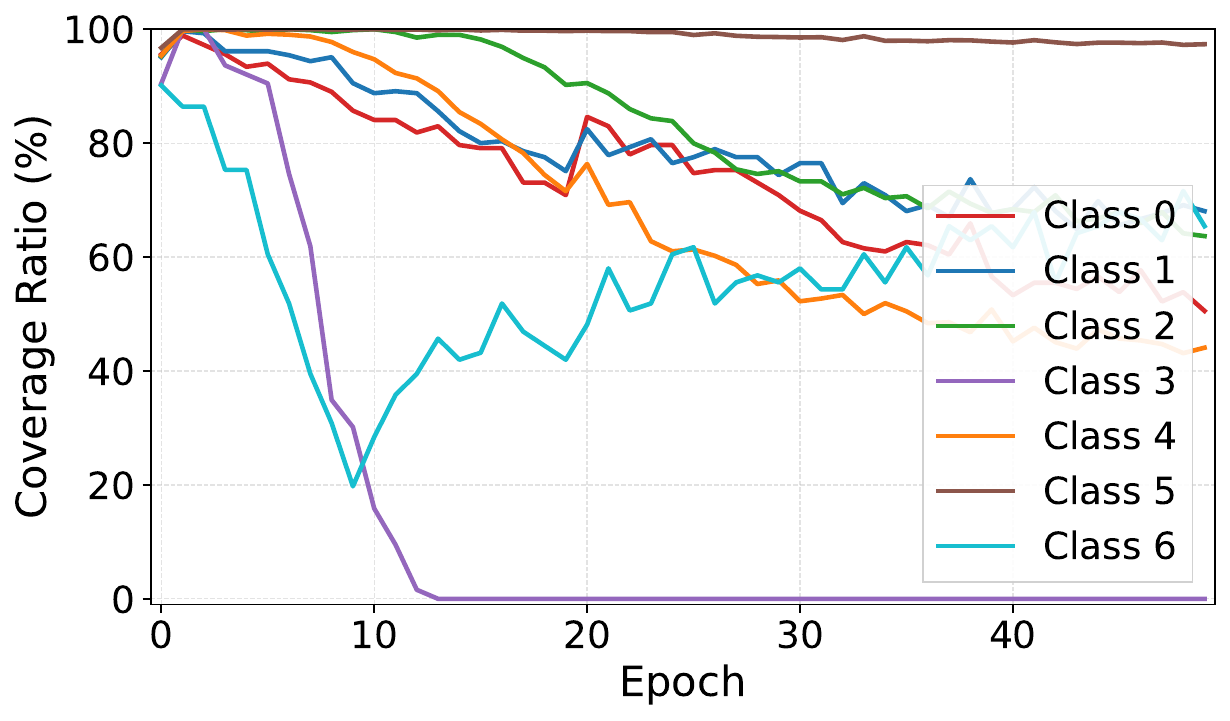}
    \caption{}
    \label{fig:recall_derdisc}
  \end{subfigure}

  \caption{Per-class performance comparison of Co-teaching and DISC methods on DermMNIST dataset with Idn-20\% noise. 
  (a) Clean ratio of Co-teaching. (b) Coverage ratio of Co-teaching. (c) Clean ratio of DISC. (d) Coverage ratio of DISC.}
  \label{fig:derma_perclass_comparison}
\end{figure}

\subsubsection{Why the Sample Selection Strategy Fails?}
Next, we further investigate the reasons for the failure of sample selection methods, using DISC and Co-teaching under Idn-20\% as examples. 

\begin{figure}
    \centering
    \includegraphics[width=0.95\columnwidth]{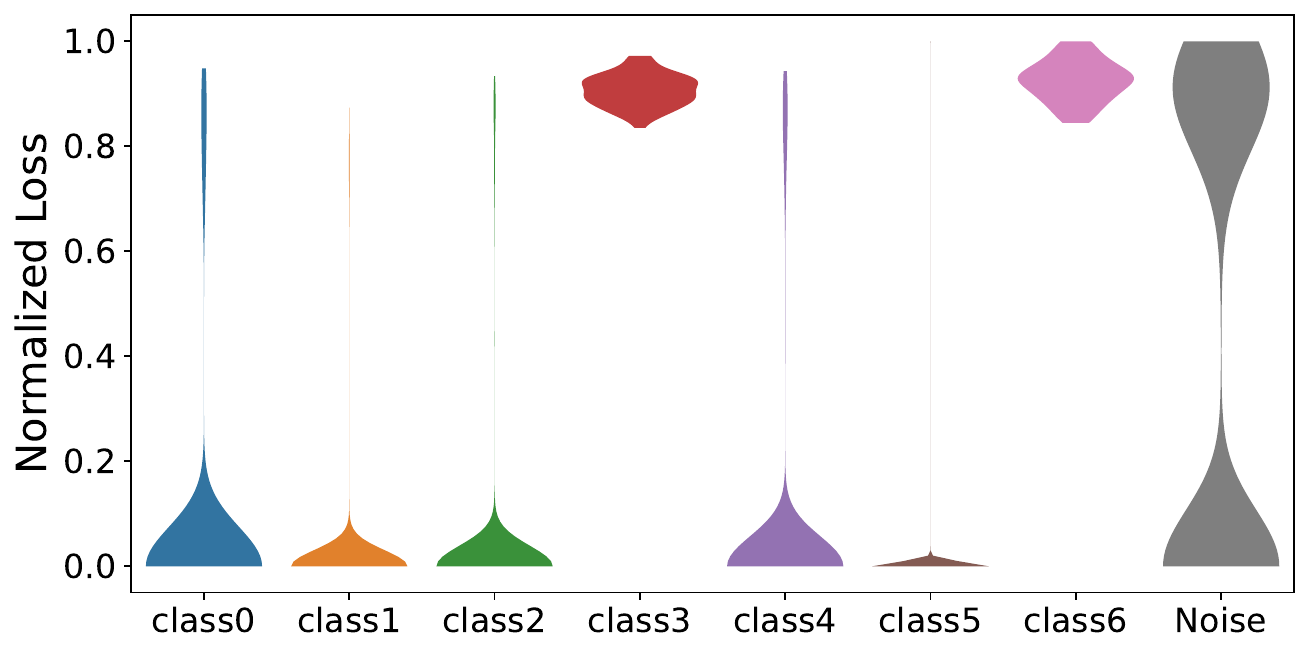} 
    \caption{
    Violin plots of per-class clean sample losses and the overall noisy sample loss distribution for Co-teaching under Idn-20\% on DermaMNIST.
    }
    \label{fig:violin}
    
\end{figure}

\textbf{Small loss strategy:} As shown in the Fig.~\ref{fig:derma_perclass_comparison}, the Co-teaching method retains almost all clean samples of class 5 for training, thereby achieving high accuracy on this class. 
In contrast, for class 3 and class 6, the method discards all clean samples at an early stage of training, resulting in zero accuracy for these classes.
For the remaining classes, most clean samples are retained for training.
However, their low clean ratios indicate that many noisy samples are also included, which leads to reduced classification accuracy.
The observed phenomenon can be explained by the distribution of losses across classes (see Fig.~\ref{fig:violin}). 
For the majority class (class 5), clean samples tend to have significantly lower losses than noisy samples. 
In contrast, for minority classes, some clean samples exhibit higher losses than noisy ones, with clean samples in classes 3 and 6 having losses exceeding those of most noisy samples.

\textbf{Dynamic instance-specific selection}: As shown in the Fig.~\ref{fig:derma_perclass_comparison}, the DISC method retains almost all clean samples of class 5 for training.
In contrast, the method discards all clean samples at an early stage of training for class 3.
For the remaining classes, although the clean ratio of the training samples is relatively high, the number of clean samples used for training decreases over time, leading to a decline in the model’s generalization capability.
This phenomenon occurs because the model fits noisy labels in the early stage of training, which lowers its confidence in clean labels and results in some clean samples being excluded from the training set.

\begin{figure}
    \centering
    \includegraphics[width=0.70\columnwidth]{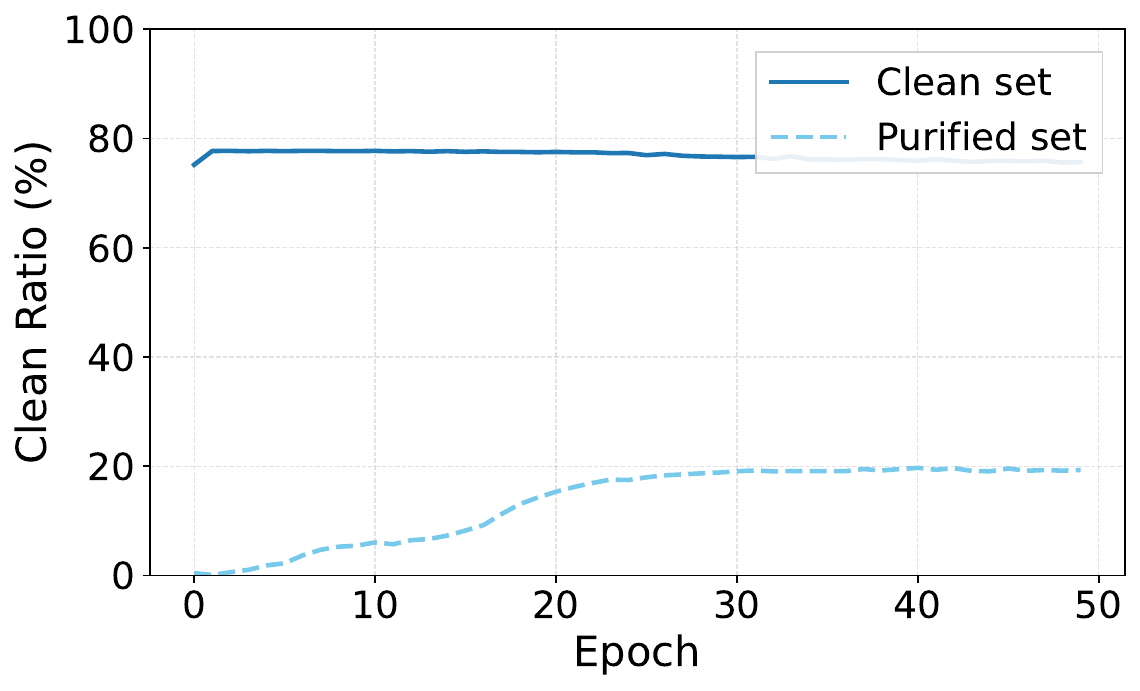} 
    \caption{
    Epoch wise ratios of clean class 5 samples in the clean set and the purified set of DISC, normalized by the total number of samples under Idn-20\% on DermaMNIST. The clean set refers to the subset of samples selected by the model as clean, while the purified set refers to the subset of excluded samples whose labels are reassigned by the model.
    }
    \label{fig:class5_trueclean_ratios}
\end{figure}

\subsubsection{Effect of the Correction Mechanism}
We also observe that, although the Co-teaching method selects almost all clean samples from class 5 for training, its accuracy on this class is only comparable to that of standard CE under Idn-20\%. 
This is because many noisy samples with low losses are selected into the clean set, including a substantial portion of class 5 samples that are incorrectly assigned noisy labels.
It is noteworthy that the DISC method achieves the highest accuracy on class 5, as it not only selects clean samples but also corrects the labels of noisy samples.
As shown in the Fig.~\ref{fig:class5_trueclean_ratios}, the clean set refers to the subset of samples selected by the model as clean, while the purified set refers to the subset of excluded samples whose labels are reassigned by the model.
We present the proportion of clean samples in the clean set and the purified set for class 5 relative to the total number of this class. 
It can be observed that nearly all samples in this class are trained with correct labels.


\begin{table}[t]
\centering
\small
\caption{Performance comparison under instance-dependent noise on ImPathMNIST. \textbf{Bold} indicates the best result.}
\label{tableImpathmnist}
\begin{tblr}{
  colspec={c c *{2}{c}}, 
  row{1} = {font=\bfseries, halign=c},
    rows = {abovesep=0.2ex, belowsep=0.2ex},  
  colsep = 3pt,      
  hline{1,2,Z} = {-}{},
  row{3-5}   = {bg=cecol},
  row{6-11}  = {bg=regucol},
  row{12-17} = {bg=samcol},
  row{18-23} = {bg=semicol},
  cells={halign=c},
}
Dataset     &      & \SetCell[c=2]{c}ImPathMNIST  \\
Noise type  &      & Idn-20\% & Idn-50\% \\
\hline
            & B    & 88.19 $\pm$ 1.25  & 70.71 $\pm$ 1.87     \\
CE          & V    & 88.19 $\pm$ 1.25   & 70.68 $\pm$ 1.92    \\
            & L    & 82.53 $\pm$ 1.43   & 55.24 $\pm$ 2.38   \\
\hline
            & B    & 88.51 $\pm$ 0.38  & 80.12 $\pm$ 0.51 \\
SCE         & V    & 88.05 $\pm$ 0.67   & 73.24  $\pm$ 0.83   \\
            & L    & 78.62 $\pm$ 2.58  & 55.34  $\pm$ 2.79   \\
\hline
            & B    & 88.01 $\pm$ 0.65  & \textbf{80.71 $\pm$ 0.98} \\
CDR         & V    & 87.76 $\pm$ 0.28   & 80.71  $\pm$ 0.72   \\
            & L    & 87.13 $\pm$ 0.43   & 60.76 $\pm$ 1.14    \\
\hline    
            & B    & 90.29 $\pm$ 0.45   & 76.34 $\pm$ 1.28     \\
Co-teaching & V    & 90.23 $\pm$ 0.63   & 75.89 $\pm$ 0.94     \\
            & L    & 87.58 $\pm$ 0.98   & 70.99 $\pm$ 1.02    \\
\hline
            & B    & 89.38 $\pm$ 0.72   & 79.88 $\pm$ 1.16     \\
CoDis       & V    & 88.69 $\pm$ 1.08   & 72.13 $\pm$ 1.09     \\
            & L    & 86.16 $\pm$ 1.02   & 71.85 $\pm$ 0.65    \\
\hline
            & B    & 91.45 $\pm$ 0.38   & 80.70 $\pm$ 0.93    \\
DISC        & V    & 91.45 $\pm$ 0.41  & \textbf{77.42 $\pm$ 1.12} \\
            & L    & \textbf{88.23 $\pm$ 1.24} & \textbf{72.04 $\pm$ 1.48} \\
\hline
            & B    & \textbf{93.25 $\pm$ 0.95} & 79.16 $\pm$ 1.52     \\
DivideMix   & V    & \textbf{93.24 $\pm$ 0.34} & 74.09 $\pm$ 0.81     \\
            & L    & 83.99 $\pm$ 1.42   & 70.40 $\pm$ 1.88     \\

\end{tblr}
\end{table}

\subsubsection{Impact of Class Imbalance on LNL}
\begin{figure}
    \centering

    \begin{subfigure}[b]{0.48\columnwidth}
        \centering
        \includegraphics[width=\columnwidth]{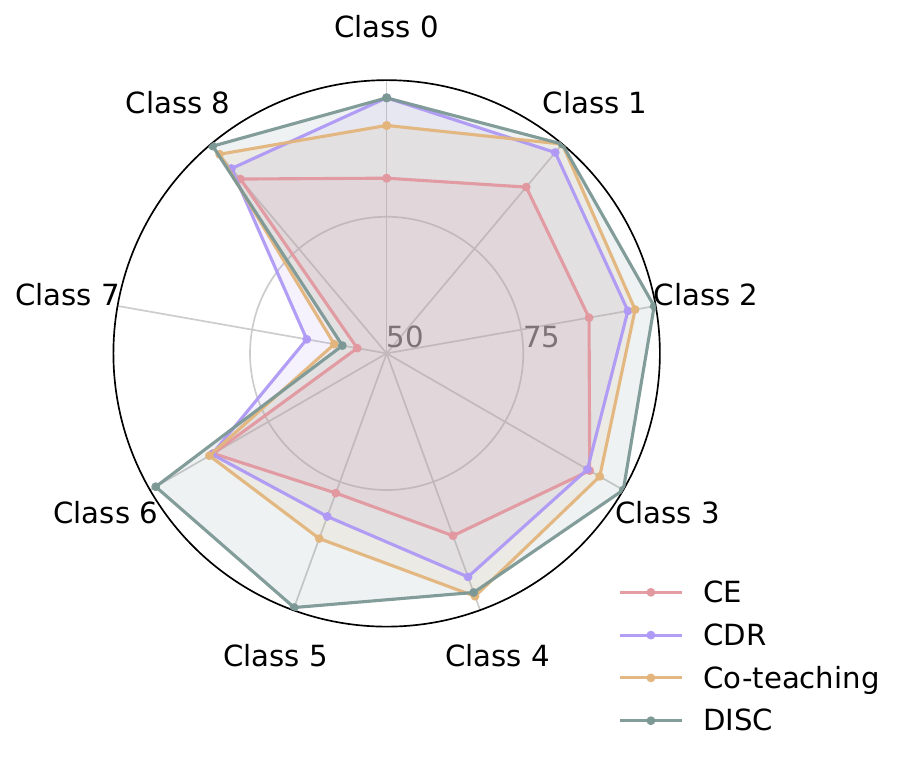}
        \caption{PathMNIST}
        \label{perclass_pathmnist}
    \end{subfigure}
    \hfill
    \begin{subfigure}[b]{0.48\columnwidth}
        \centering
        \includegraphics[width=\columnwidth]{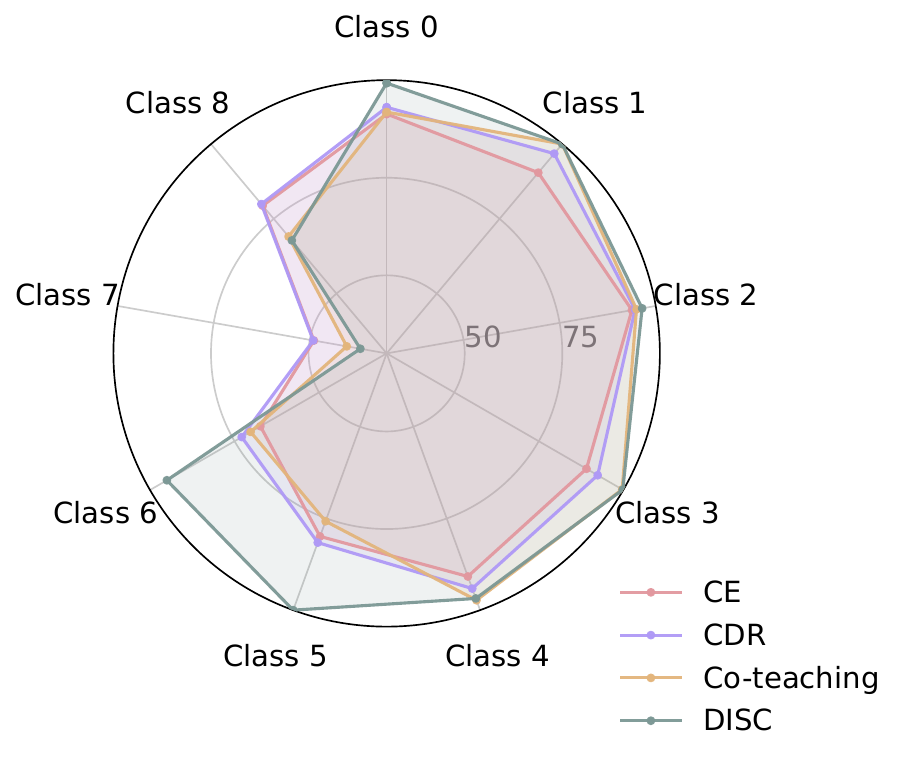}
        \caption{ImPathMNIST}
        \label{perclass_impathmnist}
    \end{subfigure}

    \caption{Per-class classification accuracy of four methods on (a) PathMNIST and (b) ImPathMNIST.}
    \label{im_vs_bl}
\end{figure}

To investigate the impact of class imbalance on LNL methods, we construct a long tailed version of the PathMNIST, denoted as ImPathMNIST. 
Following the setting in~\cite{liu2019large}, we simulate a long-tailed distribution by downsampling each class according to an exponential decay rule. 
Given $k$ classes and an imbalance ratio $r = \frac{N_0}{N_{k-1}}$, where $N_0$ is the number of samples in the most frequent (head) class, the number of samples for class $j \in [0, k)$ is calculated as:
\begin{equation}
    N_j = N_0 \cdot r^{- \frac{j}{k-1}}
\end{equation}
We use the number of samples in class 0 as $N_0$, and set $r = 100$ to create the long-tailed distribution. For each class, we randomly select the top $N_j$ samples to form the final training set. This setting allows us to systematically evaluate the robustness of label noise methods under severe class imbalance.
In this experiment, we excluded T-Revision, VolMinNet, JoCoR and Co-teaching+ due to their inferior performance.
As shown in the Table~\ref{tableImpathmnist}, all methods, except SCE, outperform CE under both Idn-20\% and Idn-50\%.
DISC achieves the best overall performance. 
As shown in the Fig.~\ref{im_vs_bl}, we compare the per-class accuracy of CE and Co-teaching, DISC and CDR under Idn-20\% with PathMNIST and ImPathMNIST.
For PathMNIST, all methods outperform CE in every class.
For ImPathMNIST, although DISC and Co-teaching achieve higher overall accuracy than CE, their performance on minority classes is inferior. 
In contrast, CDR consistently surpasses CE across all classes. 
These results indicate that the effectiveness of sample selection-based methods is sensitive to class imbalance.

\textbf{Findings:} 
Based on sample selection methods rely on the model’s loss or confidence to select clean samples. 
However, in imbalanced datasets, some clean samples from minority classes exhibit higher losses than noisy samples during the early training stage, preventing them from being selected. 
This further limits the model’s ability to learn minority classes. 
Consequently, their performance on imbalanced datasets can be even worse than that of standard CE model.
In addition, the correction mechanism enables the model to utilize noisy samples by relabeling them, leading to more complete and accurate training compared to pure sample selection methods.

\subsection{Results on Real-world Noisy Datasets}\label{realworldsec}
In this section, we evaluate the performance of LNL methods on real-world noisy datasets (see Table.~\ref{realworld}). 
Specifically, we select two retinal image datasets, DRTid and Kaggle DR+, as well as a CT image dataset, CheXpert.

\subsubsection{DRTiD}
\begin{figure*}
    \centering

    \begin{subfigure}[b]{0.32\textwidth}
        \centering
        \includegraphics[width=\columnwidth]{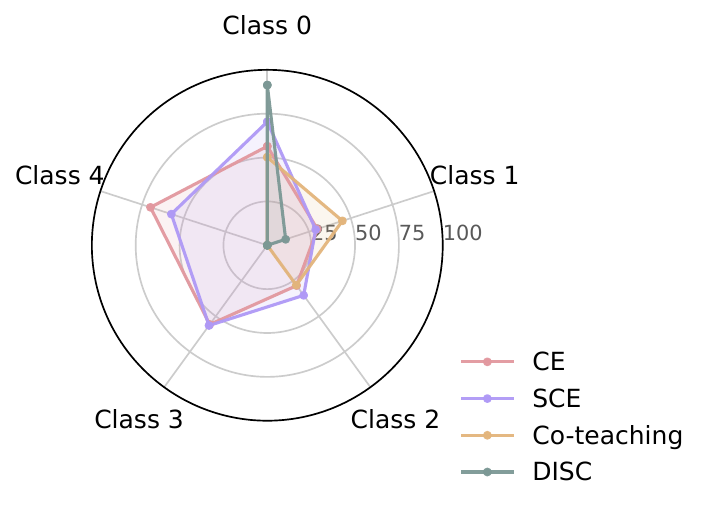}
        \caption{}
        \label{perclass_drtid}
    \end{subfigure}
    \hfill
    \begin{subfigure}[b]{0.32\textwidth}
        \centering
        \includegraphics[width=\columnwidth]{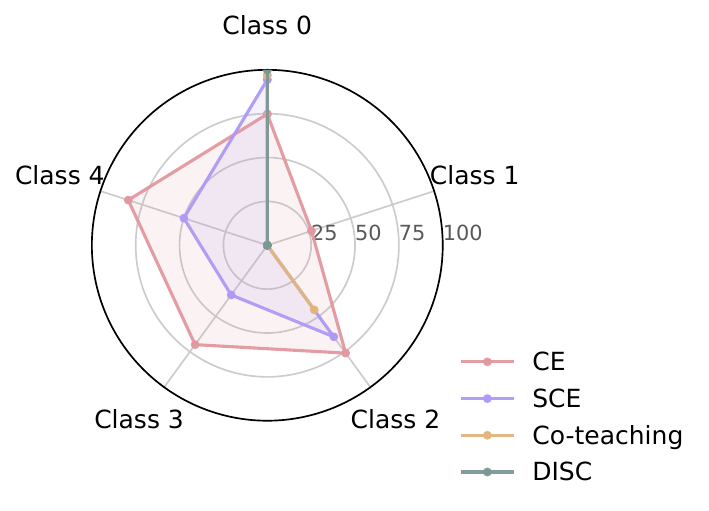}
        \caption{}
        \label{perclass_kaggledr}
    \end{subfigure}
    \hfill
    \begin{subfigure}[b]{0.32\textwidth}
        \centering
        \includegraphics[width=\columnwidth]{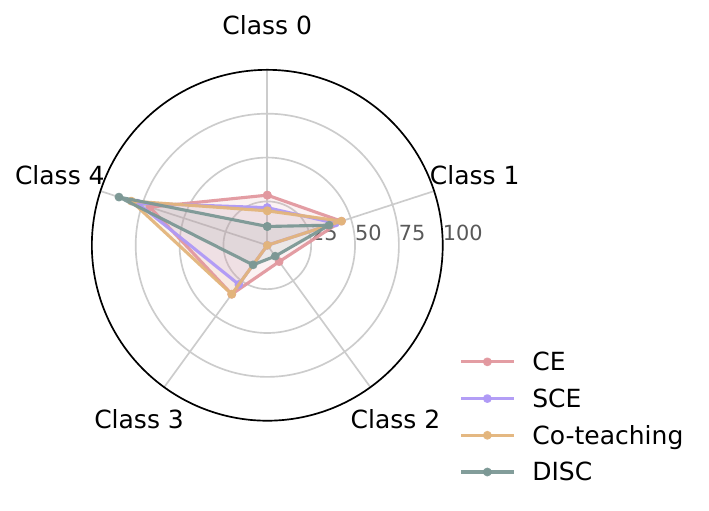}
        \caption{}
        \label{perclass_che}
    \end{subfigure}
    \caption{Per-class classification accuracy of four methods on (a) DRTiD (b) Kaggle DR+ and (c) CheXpert.}
    \label{perclassrealworld}
\end{figure*}

As shown in Table~\ref{realworld}, only SCE achieves higher performance than CE in terms of L, whereas all other methods underperform compared to CE. 
Sample selection based methods perform poorly on DRTiD, and CDR exhibits a performance gap of 15.45\% between V and L.
As shown in Fig.~\ref{perclass_drtid}, we further report the per-class accuracy of three representative methods and CE. 
SCE outperforms CE in every class except class 5. 
DISC is only effective on class 0, achieving the highest accuracy for this class while yielding near zero accuracy for all other classes. 
Co-teaching method achieves comparable accuracy to CE on the first three classes but fails completely on the last two classes. 
\begin{table}[t]
\centering
\caption{Performance comparison on real-world noisy datasets.  
\textbf{Bold} indicates the best result, 
\underline{underline} indicates the second best, 
and \uuline{double underline} indicates the third best.}
\label{realworld}
\scriptsize

\begin{tblr}{
  cell{1}{3} = {c=3}{},
  cell{3}{1}  = {r=3}{},
  cell{6}{1}  = {r=3}{},
  cell{9}{1}  = {r=3}{},
  cell{12}{1} = {r=3}{},
  cell{15}{1} = {r=3}{},
  cell{18}{1} = {r=3}{},
  cell{21}{1} = {r=3}{},
  cell{24}{1} = {r=3}{},
  cell{27}{1} = {r=3}{},
  cell{30}{1} = {r=3}{},
  cell{33}{1} = {r=3}{},
  cell{36}{1} = {r=3}{},
  cell{39}{1} = {r=3}{},
  colspec={c c *{3}{c}},
  rows  = {abovesep=0.3ex, belowsep=0.3ex},
  colsep = 3pt,
  row{1} = {font=\bfseries, halign=c},
  hline{1,2,3,Z} = {-}{},
  row{3-5}   = {bg=cecol},
  row{6-8}   = {bg=matrixcol},
  row{9-11}  = {bg=matrixcol},
  row{12-14} = {bg=regucol},
  row{15-17} = {bg=regucol},
  row{18-20} = {bg=samcol},
  row{21-23} = {bg=samcol},
  row{24-26} = {bg=samcol},
  row{27-29} = {bg=samcol},
  row{30-32} = {bg=semicol},
  row{33-35} = {bg=semicol},
  row{36-38} = {bg=semicol},
  row{39-41} = {bg=semicol},
}
 &  & \SetCell[c=3]{c}Real-world Datasets \\
 Dataset &  & DRTID & Kaggle DR+ & CheXpert \\
CE
  & B & \underline{61.46 $\pm$ 1.23} & \textbf{70.73 $\pm$ 1.42} & \uuline{43.07 $\pm$ 2.47} \\
  & V & \textbf{59.36 $\pm$ 2.12} & \uuline{67.17 $\pm$ 1.71} & 34.31 $\pm$ 2.98 \\
  & L & 46.36 $\pm$ 1.26 & 65.59 $\pm$ 1.85 & 37.08 $\pm$ 1.19 \\
  \midrule
T-Revision
  & B & 54.24 $\pm$ 1.37 & 58.42 $\pm$ 1.61 & 40.52 $\pm$ 2.31 \\
  & V & 53.63 $\pm$ 3.26 & 58.20 $\pm$ 1.35 & 35.76 $\pm$ 1.14 \\
  & L & 40.18 $\pm$ 1.98 & 57.77 $\pm$ 0.97 & 33.57 $\pm$ 1.63 \\
  \midrule
VolMinNet
  & B & 59.47 $\pm$ 0.66 & 61.62 $\pm$ 1.35 & 39.42 $\pm$ 2.58 \\
  & V & 47.73 $\pm$ 1.79 & 57.39 $\pm$ 2.47 & 39.42 $\pm$ 1.72 \\
  & L & 47.51 $\pm$ 0.77 & 54.19 $\pm$ 2.43 & 28.61 $\pm$ 1.49 \\
  \midrule
SCE
  & B & 58.85 $\pm$ 0.68 & 66.13 $\pm$ 0.81 & \textbf{48.96 $\pm$ 1.54} \\
  & V & 55.07 $\pm$ 0.76 & 65.50 $\pm$ 0.74 & \underline{38.24 $\pm$ 1.72} \\
  & L & \underline{56.01 $\pm$ 0.95} & \uuline{65.86 $\pm$ 0.79} & \underline{40.46 $\pm$ 1.48} \\
  \midrule
CDR
  & B & 54.09 $\pm$ 0.82 & \underline{70.19 $\pm$ 0.67} & 42.33 $\pm$ 1.41 \\
  & V & 52.55 $\pm$ 0.58 & \underline{67.49 $\pm$ 0.93} & 35.77 $\pm$ 1.27 \\
  & L & \uuline{50.78 $\pm$ 1.37} & \underline{67.93 $\pm$ 0.59} & \uuline{37.96 $\pm$ 2.16} \\
  \midrule
Co-teaching
  & B & 48.18 $\pm$ 1.34 & 62.90 $\pm$ 1.63 & 37.22 $\pm$ 2.18 \\
  & V & 36.86 $\pm$ 1.67 & 62.80 $\pm$ 0.74 & 35.76 $\pm$ 2.35 \\
  & L & 38.25 $\pm$ 1.25 & 62.35 $\pm$ 0.89 & 35.33 $\pm$ 2.27 \\
  \midrule
Co-teaching+
  & B & 49.31 $\pm$ 1.26 & 53.99 $\pm$ 1.31 & 28.10 $\pm$ 1.37 \\
  & V & 45.18 $\pm$ 2.74 & 53.28 $\pm$ 0.98 & 27.31 $\pm$ 1.45 \\
  & L & 46.75 $\pm$ 2.13 & 51.48 $\pm$ 2.16 & 19.71 $\pm$ 2.19 \\
  \midrule
CoDis
  & B & 49.64 $\pm$ 2.06 & 53.99 $\pm$ 2.85 & 37.72 $\pm$ 3.85 \\
  & V & 42.57 $\pm$ 5.75 & 51.03 $\pm$ 2.48 & 35.95 $\pm$ 2.26 \\
  & L & 41.80 $\pm$ 5.35 & 52.14 $\pm$ 2.38 & 34.26 $\pm$ 2.44 \\
  \midrule
JoCoR
  & B & 47.09 $\pm$ 1.25 & 64.97 $\pm$ 1.84 & 39.78 $\pm$ 1.92 \\
  & V & 44.28 $\pm$ 0.98 & 62.44 $\pm$ 1.06 & 33.94 $\pm$ 1.16 \\
  & L & 41.26 $\pm$ 1.31 & 63.13 $\pm$ 0.71 & 35.04 $\pm$ 1.65 \\
  \midrule
DISC
  & B & 48.18 $\pm$ 0.42 & 55.17 $\pm$ 0.87 & 40.15 $\pm$ 1.57 \\
  & V & 44.27 $\pm$ 0.74 & 53.88 $\pm$ 1.08 & 33.57 $\pm$ 1.11 \\
  & L & 45.22 $\pm$ 0.67 & 54.00 $\pm$ 0.61 & 31.39 $\pm$ 0.93 \\
  \midrule
DivideMix
  & B & 53.64 $\pm$ 0.73 & 67.39 $\pm$ 0.36 & 33.58 $\pm$ 1.35 \\
  & V & 52.73 $\pm$ 0.66 & 67.39 $\pm$ 0.83 & 33.58 $\pm$ 0.97 \\
  & L & 41.09 $\pm$ 0.94 & 54.00 $\pm$ 0.86 & 21.17 $\pm$ 0.71 \\
  \midrule
DivideMix+MedSSL
  & B & \uuline{59.50 $\pm$ 1.86} & 65.53 $\pm$ 1.42 & 41.06 $\pm$ 1.69 \\
  & V & \uuline{56.95 $\pm$ 1.86} & 62.73 $\pm$ 1.48 & \uuline{38.20 $\pm$ 0.54} \\
  & L & 50.73 $\pm$ 0.77 & 63.13 $\pm$ 1.95 & 35.01 $\pm$ 1.19 \\
  \midrule
DISC+MedSSL
  & B & \textbf{62.61 $\pm$ 1.94} & \uuline{69.17 $\pm$ 0.63} & \underline{44.95 $\pm$ 1.98} \\
  & V & \underline{58.24 $\pm$ 1.68} & \textbf{68.90 $\pm$ 0.95} & \textbf{41.16 $\pm$ 1.25} \\
  & L & \textbf{57.18 $\pm$ 0.79} & \textbf{68.47 $\pm$ 1.03} & \textbf{40.71 $\pm$ 1.28} \\
\end{tblr}
\end{table}

\subsubsection{Kaggle DR+}
As shown in Table~\ref{realworld}, only SCE and CDR achieve slightly better performance than CE on Kaggle DR+.
Sample selection based methods perform poorly on Kaggle DR+.
As shown in Fig.~\ref{perclass_kaggledr}, we further report the per-class accuracy of three representative methods and CE. 
All methods surpass CE only in class 0, while in other classes their performance falls behind CE.
DISC is only effective on class 0, achieving the highest accuracy for this class while yielding near zero accuracy for all other classes. 
Co-teaching achieves comparable accuracy to CE on the class 0 but fails completely on the other classes. 
Although SCE achieves better performance than the other two methods on minority classes, its performance remains inferior to that of CE.
Due to the lack of ground-truth labels, we refrain from performing further analysis on this dataset.
\subsubsection{CheXpert}
As shown in Table~\ref{realworld}, only SCE and CDR achieve slightly better performance than CE on CheXpert.
Sample selection based methods perform poorly on the CheXpert.
We further report the per-class accuracy of three representative methods and CE. 
All methods achieve the highest accuracy on class 4, which is the majority class.
As shown in Fig.~\ref{perclass_che}, the accuracy of these three methods exceeds that of CE only for class 4.
We also notice that some methods yield zero accuracy on certain classes, further suggesting that existing label noise approaches are not well suited for real-world noisy data.

\textbf{Findings:} Current LNL methods perform poorly on real-world medical datasets, in some cases even underperforming standard CE model. 
Moreover, semi-supervised methods that perform well on synthetic noisy datasets often fail to achieve comparable performance on real-world noisy medical datasets.
\subsection{Summary}\label{summarysec}
Our experiments highlight several limitations of existing LNL methods in medical imaging. 
\begin{itemize}
\item  Pure sample-selection methods offer limited practical value since they assume access to the noise rate, which is rarely available in deployment. 
Both sample selection and semi-supervised methods are further affected by class imbalance, where minority-class examples are often misclassified as noise. 
\item 
Noise-robust regularization methods are only effective under low noise levels, and transition matrix estimation methods, typically designed for class-conditional noise.
\item Existing methods fail under severe noise conditions, including high noise levels and real-world noise.
\item There is a shortage of real-world noisy datasets with reliable ground-truth labels. 
Such resources are essential to analyze why current methods fail under real-world medical noise conditions.
\end{itemize}

Therefore, future research could benefit from the following directions. 
\begin{itemize}
\item Future research should focus on designing more effective sample selection strategy to alleviate the impact of class imbalance and severe noise. 
In particular, purified label strategies hold promise, as they can not only support semi-supervised training but also serve as a means of data cleansing, which is especially valuable in medical image analysis.
\item 
Unlike anchor-based methods that produce a class-conditional transition matrix, instance-dependent transition matrix estimation approaches estimate a transition matrix for each individual sample, thereby capturing sample-specific noise characteristics.
\item 
Both methods can be further enhanced by incorporating noise-robust regularization techniques to improve stability and generalization.
\end{itemize}

\section{MedSSL}
Motivated by these findings, we next focus on improving semi-supervised methods~\cite{li2023disc,li2020dividemix} to more effectively address label noise in medical image analysis.
From the above experiments, we observe three major challenge. 
First, models tend to overfit noisy labels at warm-up stage. 
Second, the methods can inadvertently admit noisy samples into the clean set. 
Third, under class imbalance, minority-class examples, being few and harder to learn, are often misjudged as noise and excluded from the clean set.
Specifically, previous methods penalize over-confident predictions by adding a negative-entropy regularizer to the cross entropy loss~\cite{li2020dividemix} to prevent rapid overfitting to noisy labels during warm-up stage.
\begin{equation}
\mathcal{H}
= - \sum_{k=0}^{K} p(k\mid x)\,\log p(k\mid x),
\end{equation}
where $K$ denotes the number of classes and $p(k\mid x)$ is the model’s predicted probability that sample $x$ belongs to class $k$.
While the negative-entropy regularizer effectively prevents overfitting to noisy labels during warm-up, it tends to delay confident predictions for minority classes, thereby amplifying the adverse effects of class imbalance in the early stage.
Therefore, we employ label smoothing to mitigate overfitting to noisy labels in warm-up stage.
We observe that regularization methods can effectively suppress the influence of noisy labels during training. 
Therefore, we introduce Reverse Cross-Entropy (RCE)~\cite{wang2019symmetric} into the clean set objective. 
The RCE loss is defined as:
\begin{equation}
L_{rce}(x)= - \sum_{k=1}^{K} p(k\mid x)\,\log q(k\mid x),
\end{equation}
where $K$ is the number of classes; $p(k\mid x)$ is the model’s predicted probability that sample $x$ belongs to class $k$, $q(k\mid x)$ is the target label distribution for $x$.
\begin{equation}
L_c = L_{ce}+L_{rce},
\end{equation}
In semi-supervised methods, confidence-based thresholding is commonly employed to construct the clean set. 
However, since majority classes tend to achieve higher prediction confidence while minority classes exhibit lower confidence, using a unified threshold biases the clean set towards majority classes and excludes most of the minority-class samples.
Let $n_k$ be the number of samples in class $k$ ($k\in\{1,\dots,K\}$), and let $n_{\min}$ and $n_{\max}$ denote the minimum and maximum class counts, respectively. We first obtain a log-scale normalized factor:
\begin{equation}
\alpha_{k} \;=\; \frac{\log n_{k} - \log n_{\min}}{\log n_{\max} - \log n_{\min} + \varepsilon}\,, \qquad \alpha_k \in [0,1)\,,
\end{equation}
where $\varepsilon>0$ is a small constant for numerical stability. 
Given two bounds $\lambda_{\text{head}}$ (for head classes) and $\lambda_{\text{tail}}$ (for tail classes), the threshold for class $k$ is
\begin{equation}
\lambda_{k} \;=\; \lambda_{\text{head}} \;-\; \bigl(\lambda_{\text{head}}-\lambda_{\text{tail}}\bigr)\,\alpha_{k}\,.
\end{equation}

We evaluate our modifications on three real-world noisy datasets.
As shown in Table~\ref{realworld}, the modified DISC achieves the highest accuracy across all three datasets.
Although the modified DivideMix~\cite{li2020dividemix} still underperforms compared to the regularization method and standard CE, it nonetheless shows substantial improvements over its original version.
This is mainly because DivideMix~\cite{li2020dividemix} relies on the small-loss strategy to select clean samples, and our previous experiments have demonstrated that this strategy is susceptible to class imbalance.
For instance, on DRTiD~\cite{hou2022cross}, the L metric improves by 9.64\% after our modification.
Although our modifications bring improvements, the overall performance on real-world noisy medical datasets is still limited. 
This suggests that semi-supervised strategies for LNL methods in medical imaging have considerable room for further improvement.

%% file: conclusion.tex
\section{Conclusion}
We built LNMBench of LNL methods for medical image analysis and evaluated them on both synthetic and real-world datasets. 
We revealed the limitations of existing approaches, showing that they remain inadequate for real-world medical scenarios, particularly under real-world noise and high ratio noise. 
By providing these insights together with our open-source implementation, we aim to support the community and inspire future research on robust learning with noisy labels in medical imaging.
Guided by these findings, we further introduce a semi-supervised approach with minor modifications to mitigate clean set contamination and improve performance on minority classes in medical imaging.

\textbf{Limitations and Future Works} This study has several limitations. First, due to the lack of publicly available implementations, we did not include label-noise learning methods specifically designed for medical datasets in our benchmark. Second, our proposed modifications only led to marginal improvements over existing semi-supervised methods. 
Future work will expand the benchmark to include real-world noisy medical datasets, more specialized LNL methods for medical images, and diversified evaluation metrics.
We also aim to maintain LNMBench as an open, evolving platform to foster fair and reproducible research in this field.

%% file: cas-dc-sample.bbl
\begin{thebibliography}{50}
\expandafter\ifx\csname natexlab\endcsname\relax\def\natexlab#1{#1}\fi
\providecommand{\url}[1]{\texttt{#1}}
\providecommand{\href}[2]{#2}
\providecommand{\path}[1]{#1}
\providecommand{\DOIprefix}{doi:}
\providecommand{\ArXivprefix}{arXiv:}
\providecommand{\URLprefix}{URL: }
\providecommand{\Pubmedprefix}{pmid:}
\providecommand{\doi}[1]{\href{http://dx.doi.org/#1}{\path{#1}}}
\providecommand{\Pubmed}[1]{\href{pmid:#1}{\path{#1}}}
\providecommand{\bibinfo}[2]{#2}
\ifx\xfnm\relax \def\xfnm[#1]{\unskip,\space#1}\fi
\bibitem[{Chen et~al.(2023)Chen, Cheng, Du, Xu, Jiang and Wang}]{chen2023two}
\bibinfo{author}{Chen, M.}, \bibinfo{author}{Cheng, H.}, \bibinfo{author}{Du, Y.}, \bibinfo{author}{Xu, M.}, \bibinfo{author}{Jiang, W.}, \bibinfo{author}{Wang, C.}, \bibinfo{year}{2023}.
\newblock \bibinfo{title}{Two wrongs don’t make a right: Combating confirmation bias in learning with label noise}, in: \bibinfo{booktitle}{Proceedings of the AAAI Conference on Artificial Intelligence}, pp. \bibinfo{pages}{14765--14773}.
\bibitem[{Chen et~al.(2019)Chen, Liao, Chen and Zhang}]{chen2019understanding}
\bibinfo{author}{Chen, P.}, \bibinfo{author}{Liao, B.B.}, \bibinfo{author}{Chen, G.}, \bibinfo{author}{Zhang, S.}, \bibinfo{year}{2019}.
\newblock \bibinfo{title}{Understanding and utilizing deep neural networks trained with noisy labels}, in: \bibinfo{booktitle}{International conference on machine learning}, \bibinfo{organization}{PMLR}. pp. \bibinfo{pages}{1062--1070}.
\bibitem[{Cordeiro and Carneiro(2020)}]{cordeiro2020survey}
\bibinfo{author}{Cordeiro, F.R.}, \bibinfo{author}{Carneiro, G.}, \bibinfo{year}{2020}.
\newblock \bibinfo{title}{A survey on deep learning with noisy labels: How to train your model when you cannot trust on the annotations?}, in: \bibinfo{booktitle}{2020 33rd SIBGRAPI conference on graphics, patterns and images (SIBGRAPI)}, \bibinfo{organization}{IEEE}. pp. \bibinfo{pages}{9--16}.
\bibitem[{Dgani et~al.(2018)Dgani, Greenspan and Goldberger}]{dgani2018training}
\bibinfo{author}{Dgani, Y.}, \bibinfo{author}{Greenspan, H.}, \bibinfo{author}{Goldberger, J.}, \bibinfo{year}{2018}.
\newblock \bibinfo{title}{Training a neural network based on unreliable human annotation of medical images}, in: \bibinfo{booktitle}{2018 IEEE 15th International symposium on biomedical imaging (ISBI 2018)}, \bibinfo{organization}{IEEE}. pp. \bibinfo{pages}{39--42}.
\bibitem[{Dosovitskiy(2020)}]{vit}
\bibinfo{author}{Dosovitskiy, A.}, \bibinfo{year}{2020}.
\newblock \bibinfo{title}{An image is worth 16x16 words: Transformers for image recognition at scale}.
\newblock \bibinfo{journal}{arXiv preprint arXiv:2010.11929} .
\bibitem[{Gehlot et~al.(2021)Gehlot, Gupta and Gupta}]{gehlot2021cnn}
\bibinfo{author}{Gehlot, S.}, \bibinfo{author}{Gupta, A.}, \bibinfo{author}{Gupta, R.}, \bibinfo{year}{2021}.
\newblock \bibinfo{title}{A cnn-based unified framework utilizing projection loss in unison with label noise handling for multiple myeloma cancer diagnosis}.
\newblock \bibinfo{journal}{Medical Image Analysis} \bibinfo{volume}{72}, \bibinfo{pages}{102099}.
\bibitem[{Ghesu et~al.(2019)Ghesu, Georgescu, Gibson, Guendel, Kalra, Singh, Digumarthy, Grbic and Comaniciu}]{ghesu2019quantifying}
\bibinfo{author}{Ghesu, F.C.}, \bibinfo{author}{Georgescu, B.}, \bibinfo{author}{Gibson, E.}, \bibinfo{author}{Guendel, S.}, \bibinfo{author}{Kalra, M.K.}, \bibinfo{author}{Singh, R.}, \bibinfo{author}{Digumarthy, S.R.}, \bibinfo{author}{Grbic, S.}, \bibinfo{author}{Comaniciu, D.}, \bibinfo{year}{2019}.
\newblock \bibinfo{title}{Quantifying and leveraging classification uncertainty for chest radiograph assessment}, in: \bibinfo{booktitle}{International conference on medical image computing and computer-assisted intervention}, \bibinfo{organization}{Springer}. pp. \bibinfo{pages}{676--684}.
\bibitem[{Gutbrod et~al.(2025)Gutbrod, Rauber, Nunes and Palm}]{gutbrod2025openmibood}
\bibinfo{author}{Gutbrod, M.}, \bibinfo{author}{Rauber, D.}, \bibinfo{author}{Nunes, D.W.}, \bibinfo{author}{Palm, C.}, \bibinfo{year}{2025}.
\newblock \bibinfo{title}{Openmibood: Open medical imaging benchmarks for out-of-distribution detection}, in: \bibinfo{booktitle}{Proceedings of the Computer Vision and Pattern Recognition Conference}, pp. \bibinfo{pages}{25874--25886}.
\bibitem[{Han et~al.(2018)Han, Yao, Yu, Niu, Xu, Hu, Tsang and Sugiyama}]{han2018co}
\bibinfo{author}{Han, B.}, \bibinfo{author}{Yao, Q.}, \bibinfo{author}{Yu, X.}, \bibinfo{author}{Niu, G.}, \bibinfo{author}{Xu, M.}, \bibinfo{author}{Hu, W.}, \bibinfo{author}{Tsang, I.}, \bibinfo{author}{Sugiyama, M.}, \bibinfo{year}{2018}.
\newblock \bibinfo{title}{Co-teaching: Robust training of deep neural networks with extremely noisy labels}.
\newblock \bibinfo{journal}{Advances in neural information processing systems} \bibinfo{volume}{31}.
\bibitem[{He et~al.(2016)He, Zhang, Ren and Sun}]{resnet50}
\bibinfo{author}{He, K.}, \bibinfo{author}{Zhang, X.}, \bibinfo{author}{Ren, S.}, \bibinfo{author}{Sun, J.}, \bibinfo{year}{2016}.
\newblock \bibinfo{title}{Deep residual learning for image recognition}, in: \bibinfo{booktitle}{Proceedings of the IEEE conference on computer vision and pattern recognition}, pp. \bibinfo{pages}{770--778}.
\bibitem[{Hendrycks et~al.(2019)Hendrycks, Lee and Mazeika}]{hendrycks2019using}
\bibinfo{author}{Hendrycks, D.}, \bibinfo{author}{Lee, K.}, \bibinfo{author}{Mazeika, M.}, \bibinfo{year}{2019}.
\newblock \bibinfo{title}{Using pre-training can improve model robustness and uncertainty}, in: \bibinfo{booktitle}{International conference on machine learning}, \bibinfo{organization}{PMLR}. pp. \bibinfo{pages}{2712--2721}.
\bibitem[{Hou et~al.(2025)Hou, Xu, Feng and Chen}]{hou2025qmix}
\bibinfo{author}{Hou, J.}, \bibinfo{author}{Xu, J.}, \bibinfo{author}{Feng, R.}, \bibinfo{author}{Chen, H.}, \bibinfo{year}{2025}.
\newblock \bibinfo{title}{Qmix: Quality-aware learning with mixed noise for robust retinal disease diagnosis}.
\newblock \bibinfo{journal}{IEEE Transactions on Medical Imaging} .
\bibitem[{Hou et~al.(2022)Hou, Xu, Xiao, Zhao, Zhang, Zou, Lu, Xue and Feng}]{hou2022cross}
\bibinfo{author}{Hou, J.}, \bibinfo{author}{Xu, J.}, \bibinfo{author}{Xiao, F.}, \bibinfo{author}{Zhao, R.W.}, \bibinfo{author}{Zhang, Y.}, \bibinfo{author}{Zou, H.}, \bibinfo{author}{Lu, L.}, \bibinfo{author}{Xue, W.}, \bibinfo{author}{Feng, R.}, \bibinfo{year}{2022}.
\newblock \bibinfo{title}{Cross-field transformer for diabetic retinopathy grading on two-field fundus images}, in: \bibinfo{booktitle}{2022 IEEE International Conference on Bioinformatics and Biomedicine (BIBM)}, \bibinfo{organization}{IEEE Computer Society}. pp. \bibinfo{pages}{985--990}.
\bibitem[{Johnson et~al.(2019)Johnson, Pollard, Greenbaum, Lungren, Deng, Peng, Lu, Mark, Berkowitz and Horng}]{johnson2019mimic}
\bibinfo{author}{Johnson, A.E.}, \bibinfo{author}{Pollard, T.J.}, \bibinfo{author}{Greenbaum, N.R.}, \bibinfo{author}{Lungren, M.P.}, \bibinfo{author}{Deng, C.y.}, \bibinfo{author}{Peng, Y.}, \bibinfo{author}{Lu, Z.}, \bibinfo{author}{Mark, R.G.}, \bibinfo{author}{Berkowitz, S.J.}, \bibinfo{author}{Horng, S.}, \bibinfo{year}{2019}.
\newblock \bibinfo{title}{Mimic-cxr-jpg, a large publicly available database of labeled chest radiographs}.
\newblock \bibinfo{journal}{arXiv preprint arXiv:1901.07042} .
\bibitem[{Ju et~al.(2022)Ju, Wang, Wang, Mahapatra, Zhao, Zhou, Liu and Ge}]{ju2022improving}
\bibinfo{author}{Ju, L.}, \bibinfo{author}{Wang, X.}, \bibinfo{author}{Wang, L.}, \bibinfo{author}{Mahapatra, D.}, \bibinfo{author}{Zhao, X.}, \bibinfo{author}{Zhou, Q.}, \bibinfo{author}{Liu, T.}, \bibinfo{author}{Ge, Z.}, \bibinfo{year}{2022}.
\newblock \bibinfo{title}{Improving medical images classification with label noise using dual-uncertainty estimation}.
\newblock \bibinfo{journal}{IEEE transactions on medical imaging} \bibinfo{volume}{41}, \bibinfo{pages}{1533--1546}.
\bibitem[{Ju et~al.(2024)Ju, Yan, Zhou, Nan, Xing, Duan and Ge}]{ju2024monica}
\bibinfo{author}{Ju, L.}, \bibinfo{author}{Yan, S.}, \bibinfo{author}{Zhou, Y.}, \bibinfo{author}{Nan, Y.}, \bibinfo{author}{Xing, X.}, \bibinfo{author}{Duan, P.}, \bibinfo{author}{Ge, Z.}, \bibinfo{year}{2024}.
\newblock \bibinfo{title}{Monica: Benchmarking on long-tailed medical image classification}.
\newblock \bibinfo{journal}{arXiv preprint arXiv:2410.02010} .
\bibitem[{Ju et~al.(2023)Ju, Yu, Wang, Zhao, Wang, Bonnington and Ge}]{ju2023hierarchical}
\bibinfo{author}{Ju, L.}, \bibinfo{author}{Yu, Z.}, \bibinfo{author}{Wang, L.}, \bibinfo{author}{Zhao, X.}, \bibinfo{author}{Wang, X.}, \bibinfo{author}{Bonnington, P.}, \bibinfo{author}{Ge, Z.}, \bibinfo{year}{2023}.
\newblock \bibinfo{title}{Hierarchical knowledge guided learning for real-world retinal disease recognition}.
\newblock \bibinfo{journal}{IEEE Transactions on Medical Imaging} \bibinfo{volume}{43}, \bibinfo{pages}{335--350}.
\bibitem[{Karimi et~al.(2020)Karimi, Dou, Warfield and Gholipour}]{karimi2020deep}
\bibinfo{author}{Karimi, D.}, \bibinfo{author}{Dou, H.}, \bibinfo{author}{Warfield, S.K.}, \bibinfo{author}{Gholipour, A.}, \bibinfo{year}{2020}.
\newblock \bibinfo{title}{Deep learning with noisy labels: Exploring techniques and remedies in medical image analysis}.
\newblock \bibinfo{journal}{Medical image analysis} \bibinfo{volume}{65}, \bibinfo{pages}{101759}.
\bibitem[{Khanal et~al.(2023)Khanal, Bhattarai, Khanal and Linte}]{khanal2023improving}
\bibinfo{author}{Khanal, B.}, \bibinfo{author}{Bhattarai, B.}, \bibinfo{author}{Khanal, B.}, \bibinfo{author}{Linte, C.A.}, \bibinfo{year}{2023}.
\newblock \bibinfo{title}{Improving medical image classification in noisy labels using only self-supervised pretraining}, in: \bibinfo{booktitle}{MICCAI Workshop on Data Engineering in Medical Imaging}, \bibinfo{organization}{Springer}. pp. \bibinfo{pages}{78--90}.
\bibitem[{Ko et~al.(2023)Ko, Yi and Yun}]{ko2023gift}
\bibinfo{author}{Ko, J.}, \bibinfo{author}{Yi, B.}, \bibinfo{author}{Yun, S.Y.}, \bibinfo{year}{2023}.
\newblock \bibinfo{title}{A gift from label smoothing: robust training with adaptive label smoothing via auxiliary classifier under label noise}, in: \bibinfo{booktitle}{Proceedings of the AAAI Conference on Artificial Intelligence}, pp. \bibinfo{pages}{8325--8333}.
\bibitem[{Li et~al.(2023a)Li, Cao, Wang, Liu, Dou, Chen and Heng}]{li2023learning}
\bibinfo{author}{Li, J.}, \bibinfo{author}{Cao, H.}, \bibinfo{author}{Wang, J.}, \bibinfo{author}{Liu, F.}, \bibinfo{author}{Dou, Q.}, \bibinfo{author}{Chen, G.}, \bibinfo{author}{Heng, P.A.}, \bibinfo{year}{2023}a.
\newblock \bibinfo{title}{Learning robust classifier for imbalanced medical image dataset with noisy labels by minimizing invariant risk}, in: \bibinfo{booktitle}{International Conference on Medical Image Computing and Computer-Assisted Intervention}, \bibinfo{organization}{Springer}. pp. \bibinfo{pages}{306--316}.
\bibitem[{Li et~al.(2020)Li, Socher and Hoi}]{li2020dividemix}
\bibinfo{author}{Li, J.}, \bibinfo{author}{Socher, R.}, \bibinfo{author}{Hoi, S.C.}, \bibinfo{year}{2020}.
\newblock \bibinfo{title}{Dividemix: Learning with noisy labels as semi-supervised learning}.
\newblock \bibinfo{journal}{arXiv preprint arXiv:2002.07394} .
\bibitem[{Li et~al.(2021)Li, Liu, Han, Niu and Sugiyama}]{volminnet}
\bibinfo{author}{Li, X.}, \bibinfo{author}{Liu, T.}, \bibinfo{author}{Han, B.}, \bibinfo{author}{Niu, G.}, \bibinfo{author}{Sugiyama, M.}, \bibinfo{year}{2021}.
\newblock \bibinfo{title}{Provably end-to-end label-noise learning without anchor points}, in: \bibinfo{booktitle}{International conference on machine learning}, \bibinfo{organization}{PMLR}. pp. \bibinfo{pages}{6403--6413}.
\bibitem[{Li et~al.(2023b)Li, Han, Shan and Chen}]{li2023disc}
\bibinfo{author}{Li, Y.}, \bibinfo{author}{Han, H.}, \bibinfo{author}{Shan, S.}, \bibinfo{author}{Chen, X.}, \bibinfo{year}{2023}b.
\newblock \bibinfo{title}{Disc: Learning from noisy labels via dynamic instance-specific selection and correction}, in: \bibinfo{booktitle}{Proceedings of the IEEE/CVF conference on computer vision and pattern recognition}, pp. \bibinfo{pages}{24070--24079}.
\bibitem[{Liao et~al.(2025a)Liao, Hu, Xie and Xia}]{liao2025instance}
\bibinfo{author}{Liao, Z.}, \bibinfo{author}{Hu, S.}, \bibinfo{author}{Xie, Y.}, \bibinfo{author}{Xia, Y.}, \bibinfo{year}{2025}a.
\newblock \bibinfo{title}{Instance-dependent label distribution estimation for learning with label noise}.
\newblock \bibinfo{journal}{International Journal of Computer Vision} \bibinfo{volume}{133}, \bibinfo{pages}{2568--2580}.
\bibitem[{Liao et~al.(2025b)Liao, Hu, Zhang and Xia}]{liao2025unleashing}
\bibinfo{author}{Liao, Z.}, \bibinfo{author}{Hu, S.}, \bibinfo{author}{Zhang, Y.}, \bibinfo{author}{Xia, Y.}, \bibinfo{year}{2025}b.
\newblock \bibinfo{title}{Unleashing the potential of open-set noisy samples against label noise for medical image classification}.
\newblock \bibinfo{journal}{Medical Image Analysis} , \bibinfo{pages}{103702}.
\bibitem[{Lin et~al.(2024)Lin, Yao and Liu}]{lin2024learning}
\bibinfo{author}{Lin, Y.}, \bibinfo{author}{Yao, Y.}, \bibinfo{author}{Liu, T.}, \bibinfo{year}{2024}.
\newblock \bibinfo{title}{Learning the latent causal structure for modeling label noise}.
\newblock \bibinfo{journal}{Advances in Neural Information Processing Systems} \bibinfo{volume}{37}, \bibinfo{pages}{120549--120577}.
\bibitem[{Litjens et~al.(2017)Litjens, Kooi, Bejnordi, Setio, Ciompi, Ghafoorian, Van Der~Laak, Van~Ginneken and S{\'a}nchez}]{litjens2017survey}
\bibinfo{author}{Litjens, G.}, \bibinfo{author}{Kooi, T.}, \bibinfo{author}{Bejnordi, B.E.}, \bibinfo{author}{Setio, A.A.A.}, \bibinfo{author}{Ciompi, F.}, \bibinfo{author}{Ghafoorian, M.}, \bibinfo{author}{Van Der~Laak, J.A.}, \bibinfo{author}{Van~Ginneken, B.}, \bibinfo{author}{S{\'a}nchez, C.I.}, \bibinfo{year}{2017}.
\newblock \bibinfo{title}{A survey on deep learning in medical image analysis}.
\newblock \bibinfo{journal}{Medical image analysis} \bibinfo{volume}{42}, \bibinfo{pages}{60--88}.
\bibitem[{Liu et~al.(2019)Liu, Miao, Zhan, Wang, Gong and Yu}]{liu2019large}
\bibinfo{author}{Liu, Z.}, \bibinfo{author}{Miao, Z.}, \bibinfo{author}{Zhan, X.}, \bibinfo{author}{Wang, J.}, \bibinfo{author}{Gong, B.}, \bibinfo{author}{Yu, S.X.}, \bibinfo{year}{2019}.
\newblock \bibinfo{title}{Large-scale long-tailed recognition in an open world}, in: \bibinfo{booktitle}{Proceedings of the IEEE/CVF conference on computer vision and pattern recognition}, pp. \bibinfo{pages}{2537--2546}.
\bibitem[{Lukasik et~al.(2020)Lukasik, Bhojanapalli, Menon and Kumar}]{labelsmoothing}
\bibinfo{author}{Lukasik, M.}, \bibinfo{author}{Bhojanapalli, S.}, \bibinfo{author}{Menon, A.}, \bibinfo{author}{Kumar, S.}, \bibinfo{year}{2020}.
\newblock \bibinfo{title}{Does label smoothing mitigate label noise?}, in: \bibinfo{booktitle}{International Conference on Machine Learning}, \bibinfo{organization}{PMLR}. pp. \bibinfo{pages}{6448--6458}.
\bibitem[{Mehrtens et~al.(2023)Mehrtens, Kurz, Bucher and Brinker}]{mehrtens2023benchmarking}
\bibinfo{author}{Mehrtens, H.A.}, \bibinfo{author}{Kurz, A.}, \bibinfo{author}{Bucher, T.C.}, \bibinfo{author}{Brinker, T.J.}, \bibinfo{year}{2023}.
\newblock \bibinfo{title}{Benchmarking common uncertainty estimation methods with histopathological images under domain shift and label noise}.
\newblock \bibinfo{journal}{Medical image analysis} \bibinfo{volume}{89}, \bibinfo{pages}{102914}.
\bibitem[{Menze et~al.(2014)Menze, Jakab, Bauer, Kalpathy-Cramer, Farahani, Kirby, Burren, Porz, Slotboom, Wiest et~al.}]{menze2014multimodal}
\bibinfo{author}{Menze, B.H.}, \bibinfo{author}{Jakab, A.}, \bibinfo{author}{Bauer, S.}, \bibinfo{author}{Kalpathy-Cramer, J.}, \bibinfo{author}{Farahani, K.}, \bibinfo{author}{Kirby, J.}, \bibinfo{author}{Burren, Y.}, \bibinfo{author}{Porz, N.}, \bibinfo{author}{Slotboom, J.}, \bibinfo{author}{Wiest, R.}, et~al., \bibinfo{year}{2014}.
\newblock \bibinfo{title}{The multimodal brain tumor image segmentation benchmark (brats)}.
\newblock \bibinfo{journal}{IEEE transactions on medical imaging} \bibinfo{volume}{34}, \bibinfo{pages}{1993--2024}.
\bibitem[{Nguyen et~al.(2019)Nguyen, Mummadi, Ngo, Nguyen, Beggel and Brox}]{nguyen2019self}
\bibinfo{author}{Nguyen, D.T.}, \bibinfo{author}{Mummadi, C.K.}, \bibinfo{author}{Ngo, T.P.N.}, \bibinfo{author}{Nguyen, T.H.P.}, \bibinfo{author}{Beggel, L.}, \bibinfo{author}{Brox, T.}, \bibinfo{year}{2019}.
\newblock \bibinfo{title}{Self: Learning to filter noisy labels with self-ensembling}.
\newblock \bibinfo{journal}{arXiv preprint arXiv:1910.01842} .
\bibitem[{Pham et~al.(2021)Pham, Le, Tran, Ngo and Nguyen}]{pham2021interpreting}
\bibinfo{author}{Pham, H.H.}, \bibinfo{author}{Le, T.T.}, \bibinfo{author}{Tran, D.Q.}, \bibinfo{author}{Ngo, D.T.}, \bibinfo{author}{Nguyen, H.Q.}, \bibinfo{year}{2021}.
\newblock \bibinfo{title}{Interpreting chest x-rays via cnns that exploit hierarchical disease dependencies and uncertainty labels}.
\newblock \bibinfo{journal}{Neurocomputing} \bibinfo{volume}{437}, \bibinfo{pages}{186--194}.
\bibitem[{Shi et~al.(2024)Shi, Zhang, Guo, Yang, Xu and Wu}]{shi2024survey}
\bibinfo{author}{Shi, J.}, \bibinfo{author}{Zhang, K.}, \bibinfo{author}{Guo, C.}, \bibinfo{author}{Yang, Y.}, \bibinfo{author}{Xu, Y.}, \bibinfo{author}{Wu, J.}, \bibinfo{year}{2024}.
\newblock \bibinfo{title}{A survey of label-noise deep learning for medical image analysis}.
\newblock \bibinfo{journal}{Medical image analysis} \bibinfo{volume}{95}, \bibinfo{pages}{103166}.
\bibitem[{Shin et~al.(2016)Shin, Roth, Gao, Lu, Xu, Nogues, Yao, Mollura and Summers}]{shin2016deep}
\bibinfo{author}{Shin, H.C.}, \bibinfo{author}{Roth, H.R.}, \bibinfo{author}{Gao, M.}, \bibinfo{author}{Lu, L.}, \bibinfo{author}{Xu, Z.}, \bibinfo{author}{Nogues, I.}, \bibinfo{author}{Yao, J.}, \bibinfo{author}{Mollura, D.}, \bibinfo{author}{Summers, R.M.}, \bibinfo{year}{2016}.
\newblock \bibinfo{title}{Deep convolutional neural networks for computer-aided detection: Cnn architectures, dataset characteristics and transfer learning}.
\newblock \bibinfo{journal}{IEEE transactions on medical imaging} \bibinfo{volume}{35}, \bibinfo{pages}{1285--1298}.
\bibitem[{Sukhbaatar et~al.(2014)Sukhbaatar, Bruna, Paluri, Bourdev and Fergus}]{sukhbaatar2014training}
\bibinfo{author}{Sukhbaatar, S.}, \bibinfo{author}{Bruna, J.}, \bibinfo{author}{Paluri, M.}, \bibinfo{author}{Bourdev, L.}, \bibinfo{author}{Fergus, R.}, \bibinfo{year}{2014}.
\newblock \bibinfo{title}{Training convolutional networks with noisy labels}.
\newblock \bibinfo{journal}{arXiv preprint arXiv:1406.2080} .
\bibitem[{Tanno et~al.(2019)Tanno, Saeedi, Sankaranarayanan, Alexander and Silberman}]{tanno2019learning}
\bibinfo{author}{Tanno, R.}, \bibinfo{author}{Saeedi, A.}, \bibinfo{author}{Sankaranarayanan, S.}, \bibinfo{author}{Alexander, D.C.}, \bibinfo{author}{Silberman, N.}, \bibinfo{year}{2019}.
\newblock \bibinfo{title}{Learning from noisy labels by regularized estimation of annotator confusion}, in: \bibinfo{booktitle}{Proceedings of the IEEE/CVF conference on computer vision and pattern recognition}, pp. \bibinfo{pages}{11244--11253}.
\bibitem[{Wang et~al.(2017)Wang, Peng, Lu, Lu, Bagheri and Summers}]{wang2017chestx}
\bibinfo{author}{Wang, X.}, \bibinfo{author}{Peng, Y.}, \bibinfo{author}{Lu, L.}, \bibinfo{author}{Lu, Z.}, \bibinfo{author}{Bagheri, M.}, \bibinfo{author}{Summers, R.M.}, \bibinfo{year}{2017}.
\newblock \bibinfo{title}{Chestx-ray8: Hospital-scale chest x-ray database and benchmarks on weakly-supervised classification and localization of common thorax diseases}, in: \bibinfo{booktitle}{Proceedings of the IEEE conference on computer vision and pattern recognition}, pp. \bibinfo{pages}{2097--2106}.
\bibitem[{Wang et~al.(2019)Wang, Ma, Chen, Luo, Yi and Bailey}]{wang2019symmetric}
\bibinfo{author}{Wang, Y.}, \bibinfo{author}{Ma, X.}, \bibinfo{author}{Chen, Z.}, \bibinfo{author}{Luo, Y.}, \bibinfo{author}{Yi, J.}, \bibinfo{author}{Bailey, J.}, \bibinfo{year}{2019}.
\newblock \bibinfo{title}{Symmetric cross entropy for robust learning with noisy labels}, in: \bibinfo{booktitle}{Proceedings of the IEEE/CVF international conference on computer vision}, pp. \bibinfo{pages}{322--330}.
\bibitem[{Wei et~al.(2020)Wei, Feng, Chen and An}]{wei2020combating}
\bibinfo{author}{Wei, H.}, \bibinfo{author}{Feng, L.}, \bibinfo{author}{Chen, X.}, \bibinfo{author}{An, B.}, \bibinfo{year}{2020}.
\newblock \bibinfo{title}{Combating noisy labels by agreement: A joint training method with co-regularization}, in: \bibinfo{booktitle}{Proceedings of the IEEE/CVF conference on computer vision and pattern recognition}, pp. \bibinfo{pages}{13726--13735}.
\bibitem[{Wei et~al.(2021)Wei, Zhu, Cheng, Liu, Niu and Liu}]{wei2021learning}
\bibinfo{author}{Wei, J.}, \bibinfo{author}{Zhu, Z.}, \bibinfo{author}{Cheng, H.}, \bibinfo{author}{Liu, T.}, \bibinfo{author}{Niu, G.}, \bibinfo{author}{Liu, Y.}, \bibinfo{year}{2021}.
\newblock \bibinfo{title}{Learning with noisy labels revisited: A study using real-world human annotations}.
\newblock \bibinfo{journal}{arXiv preprint arXiv:2110.12088} .
\bibitem[{Xia et~al.(2023)Xia, Han, Zhan, Yu, Gong, Gong and Liu}]{xia2023combating}
\bibinfo{author}{Xia, X.}, \bibinfo{author}{Han, B.}, \bibinfo{author}{Zhan, Y.}, \bibinfo{author}{Yu, J.}, \bibinfo{author}{Gong, M.}, \bibinfo{author}{Gong, C.}, \bibinfo{author}{Liu, T.}, \bibinfo{year}{2023}.
\newblock \bibinfo{title}{Combating noisy labels with sample selection by mining high-discrepancy examples}, in: \bibinfo{booktitle}{Proceedings of the IEEE/CVF international conference on computer vision}, pp. \bibinfo{pages}{1833--1843}.
\bibitem[{Xia et~al.(2020a)Xia, Liu, Han, Gong, Wang, Ge and Chang}]{xia2020robust}
\bibinfo{author}{Xia, X.}, \bibinfo{author}{Liu, T.}, \bibinfo{author}{Han, B.}, \bibinfo{author}{Gong, C.}, \bibinfo{author}{Wang, N.}, \bibinfo{author}{Ge, Z.}, \bibinfo{author}{Chang, Y.}, \bibinfo{year}{2020}a.
\newblock \bibinfo{title}{Robust early-learning: Hindering the memorization of noisy labels}, in: \bibinfo{booktitle}{International conference on learning representations}.
\bibitem[{Xia et~al.(2020b)Xia, Liu, Han, Wang, Gong, Liu, Niu, Tao and Sugiyama}]{xia2020part}
\bibinfo{author}{Xia, X.}, \bibinfo{author}{Liu, T.}, \bibinfo{author}{Han, B.}, \bibinfo{author}{Wang, N.}, \bibinfo{author}{Gong, M.}, \bibinfo{author}{Liu, H.}, \bibinfo{author}{Niu, G.}, \bibinfo{author}{Tao, D.}, \bibinfo{author}{Sugiyama, M.}, \bibinfo{year}{2020}b.
\newblock \bibinfo{title}{Part-dependent label noise: Towards instance-dependent label noise}.
\newblock \bibinfo{journal}{Advances in neural information processing systems} \bibinfo{volume}{33}, \bibinfo{pages}{7597--7610}.
\bibitem[{Xia et~al.(2019)Xia, Liu, Wang, Han, Gong, Niu and Sugiyama}]{xia2019anchor}
\bibinfo{author}{Xia, X.}, \bibinfo{author}{Liu, T.}, \bibinfo{author}{Wang, N.}, \bibinfo{author}{Han, B.}, \bibinfo{author}{Gong, C.}, \bibinfo{author}{Niu, G.}, \bibinfo{author}{Sugiyama, M.}, \bibinfo{year}{2019}.
\newblock \bibinfo{title}{Are anchor points really indispensable in label-noise learning?}
\newblock \bibinfo{journal}{Advances in neural information processing systems} \bibinfo{volume}{32}.
\bibitem[{Xue et~al.(2019)Xue, Dou, Shi, Chen and Heng}]{xue2019robust}
\bibinfo{author}{Xue, C.}, \bibinfo{author}{Dou, Q.}, \bibinfo{author}{Shi, X.}, \bibinfo{author}{Chen, H.}, \bibinfo{author}{Heng, P.A.}, \bibinfo{year}{2019}.
\newblock \bibinfo{title}{Robust learning at noisy labeled medical images: Applied to skin lesion classification}, in: \bibinfo{booktitle}{2019 IEEE 16th International symposium on biomedical imaging (ISBI 2019)}, \bibinfo{organization}{IEEE}. pp. \bibinfo{pages}{1280--1283}.
\bibitem[{Yang et~al.(2023)Yang, Shi, Wei, Liu, Zhao, Ke, Pfister and Ni}]{medmnistv2}
\bibinfo{author}{Yang, J.}, \bibinfo{author}{Shi, R.}, \bibinfo{author}{Wei, D.}, \bibinfo{author}{Liu, Z.}, \bibinfo{author}{Zhao, L.}, \bibinfo{author}{Ke, B.}, \bibinfo{author}{Pfister, H.}, \bibinfo{author}{Ni, B.}, \bibinfo{year}{2023}.
\newblock \bibinfo{title}{Medmnist v2-a large-scale lightweight benchmark for 2d and 3d biomedical image classification}.
\newblock \bibinfo{journal}{Scientific Data} \bibinfo{volume}{10}, \bibinfo{pages}{41}.
\bibitem[{Yu et~al.(2019)Yu, Han, Yao, Niu, Tsang and Sugiyama}]{yu2019does}
\bibinfo{author}{Yu, X.}, \bibinfo{author}{Han, B.}, \bibinfo{author}{Yao, J.}, \bibinfo{author}{Niu, G.}, \bibinfo{author}{Tsang, I.}, \bibinfo{author}{Sugiyama, M.}, \bibinfo{year}{2019}.
\newblock \bibinfo{title}{How does disagreement help generalization against label corruption?}, in: \bibinfo{booktitle}{International conference on machine learning}, \bibinfo{organization}{PMLR}. pp. \bibinfo{pages}{7164--7173}.
\bibitem[{Zhou et~al.(2020)Zhou, Wang and Bilmes}]{zhou2020robust}
\bibinfo{author}{Zhou, T.}, \bibinfo{author}{Wang, S.}, \bibinfo{author}{Bilmes, J.}, \bibinfo{year}{2020}.
\newblock \bibinfo{title}{Robust curriculum learning: from clean label detection to noisy label self-correction}, in: \bibinfo{booktitle}{International conference on learning representations}.

\end{thebibliography}
